\documentclass[journal, compsoc]{IEEEtran}

\usepackage[utf8]{inputenc}
\usepackage{cite}
\usepackage{graphicx}
\usepackage{tikz}
\usepackage{pgfplots}
\pgfplotsset{compat=newest,legend style={font=\footnotesize},
             ticklabel style={font=\footnotesize},
             x label style={font=\footnotesize},
             y label style={font=\footnotesize}}
\usepgfplotslibrary{groupplots,dateplot}
\usetikzlibrary{patterns,shapes.arrows}

\usepackage{comment}
\usepackage{amsmath,amssymb,amsthm,mathtools,amsfonts}
\usepackage{bm}
\usepackage{algorithm,algorithmic}
\usepackage{bbm}
\usepackage{url}
\usepackage{array}
\usepackage{xr-hyper}
\usepackage{hyperref}
\usepackage{enumerate}
\usepackage[squaren]{SIunits}
\usepackage{xspace}
\usepackage[]{hyperref}

\usepackage[commandnameprefix=ifneeded]{changes}
\makeatletter
\@namedef{Changes@AuthorColor}{magenta}
\colorlet{Changes@Color}{magenta}
\makeatother

\makeatletter
\renewcommand{\todo}[2][]{\tikzexternaldisable\@todo[#1]{#2}\tikzexternalenable}
\makeatother

\newcommand{\footlabel}[2]{%
    \addtocounter{footnote}{1}%
    \footnotetext[\thefootnote]{%
        \addtocounter{footnote}{-1}%
        \refstepcounter{footnote}\label{#1}%
        #2%
    }%
    $^{\ref{#1}}$%
}

\newcommand{\listintzero}[1]{\lbrace 0, 1, \dots, #1\rbrace}

\def\mm{\milli\metre}

\def\qinq{{q \in \Q}}

\def\dens{\tilde{\pi}}
\def\density{\dens}

\def\densityu{\pi} 
\def\densitye{\hat{\densityu}}

\newcommand{\de}{DE\xspace}

\newcommand\tc[1]{\textcolor{black}{#1}}

\def\b{\mathbf{b}}

\def\ker{b}
\def\Sensi{\mathbb{E}}
\def\sensi{\xi}
\def\sensiInt{\Xi}
\def\point{\mathbf{x}}
\def\sinodomain{\mathbb{X}}
\def\Samples{\mathbb{X}_N}
\def\SamplesCard{N}
\def\sample{\bm{x}}
\def\SamplesSub{N}
\def\qinq{{\sample \in \Samples }}

\def\func{J}
\def\lipsbest{\text{Lip}\{\mathbf{g}\}}
\def\lips{B_{\text{Lip}}}
\def\coeffs{c}
\def\coeffsm{\coeffs[\m]}
\def\opticoeffs{\coeffs^\star}
\def\indexk{\mathbf{k}}
\def\indexl{\mathbf{n}}
\def\index{\mathbf{m}}
\def\vmu{\boldsymbol{\mu}}
\def\vbeta{\varphi}

\def\test{\tilde{\coeffs}}
\def\testm{\tilde{\coeffs}[\m]}
\def\hess{\mathrm{\mathbf{H}}}
\def\samp{\mathcal{S}}

\def\vmu{\boldsymbol{\mu}}

\def\k{\mathbf{k}}
\def\l{\mathbf{l}}

\def\m{\mathbf{m}}

\def\obs{\nu} 

\DeclareMathOperator{\prox}{prox}

\DeclarePairedDelimiterX{\inp}[2]{\langle}{\rangle}{#1, #2}
\newcommand{\norm}[1]{\left\lVert#1\right\rVert}

\newcommand{\twodots}{\mathinner {\ldotp \ldotp}}
\DeclarePairedDelimiterX{\inner}[2]{\langle}{\rangle}{#1, #2}
\def\reals{\mathbb{R}}
\def\integers{\mathbb{Z}}
\def\naturals{\mathbb{N}}

\def\define{=}

\DeclareMathOperator{\ff}{\textit{b}\MRkern \textit{b}}
\newcommand{\MRkern}{%
  \mkern-1.9mu
  \mathchoice{}{}{\mkern0.3mu}{\mkern0.5mu}%
}

\DeclareMathOperator{\tr}{tr}

\def\pngpath{figs/results}
\def\tikzpath{figs}
\def\tikzpathh{figs/phantoms}

\begin{document}
\bstctlcite{IEEEexample:BSTcontrol}

\title{Sensitivity-Aware Density Estimation \\ in Multiple Dimensions} 

\author{Aleix Boquet-Pujadas
\thanks{
Corresponding author: aleix.boquetipujadas@epfl.ch
\\
The authors are with the Biomedical Imaging Group at the École polytechnique fédérale de Lausanne, 
Lausanne, Switzerland.}, Pol {del Aguila Pla},~\IEEEmembership{Member,~IEEE,}%
\thanks{Pol~{del Aguila Pla} is also with the CIBM Center for Biomedical Imaging, in Switzerland.}
and Michael Unser,~\IEEEmembership{Life Fellow,~IEEE}%
}

\maketitle

\begin{abstract}
We formulate an optimization problem to estimate probability densities in the context of multidimensional problems that are sampled with uneven probability. It considers detector sensitivity as an heterogeneous density and takes advantage of the computational speed and flexible boundary conditions offered by splines on a grid. We choose to regularize the Hessian of the spline via the nuclear norm to promote sparsity.
As a result, the method is spatially adaptive and stable against the choice of the regularization parameter, which plays the role of the bandwidth. 
We test our computational pipeline on standard densities and provide software. We also present a new approach to PET rebinning as an application of our framework. 

\end{abstract}


\section{Introduction}

        More and more imaging modalities are entering the regime of low photon counts and thus require statistical consideration.
    This is is the case of super-resolution microscopy and, more recently, emission tomography \cite{ihsani_kernel_2016,shopa_application_2018, iacobucci_monolithic_2021, pawlak_density_2005, khater_review_2020}.
    Yet, in most density estimation (DE) methods, dimensionality issues arise as early as 2D or 3D.
    As the dimension $d$ of a domain $\sinodomain \subset \reals^d$ increases, it also becomes more cumbersome to sample space evenly. Events become not only scarcer, but the consistency of the probability of detection also worsens across sensor elements.
    In some tomographic modalities, for example, sensors have their own probability of detecting photons, which can vary considerably across the scanner~\cite{iacobucci_efficiency_2022}.
    These considerations are usually incorporated into \de routines via a point-wise weighting of the measurements~\cite{gisbert_weighted_2003}.
    However, zones of low sensitivity remain problematic because the use of weights introduces stability issues upon inversion of the detection probability \cite{gisbert_weighted_2003, wolters_practical_2018}. Boundary conditions (BCs) are another consideration that is often overlooked by \de methods, yet are important to describe domains such as those of periodic sinograms.

    Other fields where these considerations are important for the application of DE include epidemiology, cellphone queries, and the study of natural phenomena \cite{doucet_distinct_2020, bourscheidt_improvements_2014, huo_short-term_2022, shi_extended_2021}. They are also afflicted by sensitivity issues, namely: diagnostic capacity, cellphone reception, and the range of weather stations. And they involve BCS too, for example to parameterize coordinates on Earth.
    
    In this paper, we develop a DE method that addresses these concerns into a unified computational framework.
    \begin{itemize}
        \item We reformulate weighted DE to avoid unstable inversions. We do it by incorporating the sensitivity as a probability-density function (pdf) that characterizes the detector~system. This corresponds to the choice of a measure.
        
        \item We express the density as an exponential family of cardinal splines in multiple dimensions. Computations scale well because the existence of an underlying grid helps us express many operations as separable convolutions. 
        
        \item We care explicitly about BCs. The spline expansion allows us to tailor the BCs to the multidimensional domain under consideration.

        \item We approach bandwidth selection by embedding the likelihood into a proximal-optimization framework. Our regularization is based on the nuclear norm of the Hessian of the underlying spline. This allows us to control the sparsity of the knots of piecewise-linear splines, along a single axis. In higher dimension,
        it favors splines that are locally affine. The effect is invariant to rotation, translation, and scale.
    \end{itemize}
    
    In comparison with standard DE methods, our experiments suggest that our framework copes better with inhomogeneous sensitivities, is computationally independent of the number of samples in higher dimension, and adapts to BCs seamlessly. They also attest to the robustness of our method to the bandwidth parameter.

   We focus on positron-emission tomography (PET) as an illustrative application. Our investigation is motivated by the original observation that new scanners turn sinogram data into point clouds that originate from an underlying Poisson process (Figure~\ref{fig:pet}).
    We show that our \de approach is adequate for PET rebinning and reconstruction in a state-of-the-art scanner.
    
    The article is divided as follows: after a review of the state of the art in Section \ref{sec:state}, we present the formulation of our framework in Section \ref{sec:proposed} and the optimization thereof in Section \ref{sec:opti}. We test the framework on standard densities in Section \ref{sec:experiments}. Section \ref{sec:applications} contains the motivation for applications, especially imaging ones, and tests of the framework on PET~rebinning.

    \section{State-of-the-art in Density Estimation} \label{sec:state}
        
        Along with histogram-based estimation (HE), kernel \de (KDE) is the most prominent among nonparametric methods for \de \cite{izenman_recent_1991}. KDE and HE are popular because of their simplicity and strong theoretical guarantees.
        That most programming libraries  implement no other method is a further testament to their ubiquity \cite{virtanen_scipy_2020}. 
        We count four main challenges to DE: dimensionality; weights; boundaries; and bandwidth.

        \subsection{Kernel Density Estimation (KDE)}

        \subsubsection{Dimensionality} KDE is statistically strongly consistent and is asymptotically normal under relatively mild assumptions. (Convergence slows down as $d$ increases \cite{izenman_recent_1991}.)
        KDE comes at a high computational cost: the evaluation of the density at a single point requires as many operations as there are samples, which parallelizes poorly. There is still ongoing work to accelerate KDE
        \cite{bullmann_fast_2018, obrien_fast_2016, elgammal_efficient_2003}. 
        One approach is to truncate the sum of kernels \cite{gray_nonparametric_2003}. (Choosing the kernels does not scale with dimension, however.)  
        Another one is to project the samples onto a grid, which gives access to convolution-based methods
        \cite{silverman_kernel_1982}. The tradeoff of many accelerated approaches is often to the detriment of accuracy.
        
        \subsubsection{Weights} A weighted KDE estimator \cite{gisbert_weighted_2003, wolters_practical_2018} for a finite set $\{\point_k\} \subset \reals^d$ of samples can be written as
        \begin{equation}\label{eq:kde}
            \frac{1}{\sum_k w_k}\sum_k w_k K_h(\point-\point_k)
        \end{equation}
        for $\point \in \reals^d$, where $\{w_k\} \subset \reals_{\geq 0}$ are the weights of the samples, and $K_h$ a kernel function with (band)width $h$. The purpose of weights is to compensate for the sensitivity. They are typically set as $w_k=1/p_k$, where $p_k$ is the probability of detection at $\point_k$. This is unsatisfying because the sensitivity is merely inverted and considered only at the sample points instead of throughout the support of the kernels.
        
        \subsubsection{Boundaries} The existence of BCs and compact domains introduce bias in KDE. Mitigations have been proposed that transform compact intervals into the real line, that extend the data past the boundary, or that use customized kernels \cite{geenens_probit_2014}. 

        \subsubsection{Bandwidth} Bandwidth selection plays an important role in KDE and remains a topic of 
        \tc{research}~\cite{izenman_recent_1991, henderson_bandwidth_2023, jiang_bandwidth_2020,kim_bandwidth_2020,tenreiro_kernel_2022,heidenreich_bandwidth_2013,terrell_variable_1992}. 
        Some \tc{(``rule-of-thumb'')} methods derive optimal bandwidths given a set of samples, for example by minimizing the mean-squared error under a assumptions such as normality, \tc{among others} \cite{heidenreich_bandwidth_2013,henderson_bandwidth_2023}; they can be made to consider the weights too. In some other \tc{(``plug-in'')} approaches, the width of the kernel is adapted locally to the samples at the expense of increased computational complexity~\cite{terrell_variable_1992,jiang_bandwidth_2020}.
        
        \vspace{-5pt}

        \subsection{Histogram Estimation (HE)}
        Although the speed of HE scales well with dimension, hyper-bins quickly become empty due to the curse of dimensionality. Precisely, the challenge of bandwidth selection appears in HE under the guise of the choice of bin size. Histograms are thus particularly sensitive to bandwidth because bins are not only discrete but also quantized. As a result, HE methods have slow convergence rates~\cite{izenman_recent_1991}. Similarly to KDE, weights are incorporated to HE via an inversion of the detection probability.
        
        \subsection{Other Methods}
         Another approach to \de is to express the estimator as an orthogonal series~\cite{schwartz_estimation_1967}. There, one adjusts the bandwidth by truncating the series expansion, which results in smoother estimates.
        In the method of logsplines, one expands the logarithm of the density and represents it as a spline defined on a tentative sequence of free knots in the real line \cite{barron_approximation_1991, kooperberg_study_1991}.
        The coefficients of the spline are then optimized to maximize the likelihood \cite{stone_large-sample_1990}.
        The bandwidth is adjusted by choosing the number and location of the knots in a way that is optimal according to criteria such as Akaike's information criterion \cite{kooperberg_comparison_2004, truong_chapter_2005}.
        One approach is to delete knots heuristically once the coefficients are fixed. Another is to optimize knots and parameters together, but the resulting minimization problem is severely non-convex. 
        A less common approach is to adopt smoothing splines to regularize the problem \cite{gu_smoothing_1993}.
        
        While the results of logsplines are promising in 1D, 
        only a couple of works extend them to 2D, where the heuristics of free knots become complicated and the implementation is much more involved than for KDEs \cite{kooperberg_bivariate_1998}. Furthermore, they do not consider weights or BCs.
        
\begin{figure}
           \centering
           \input{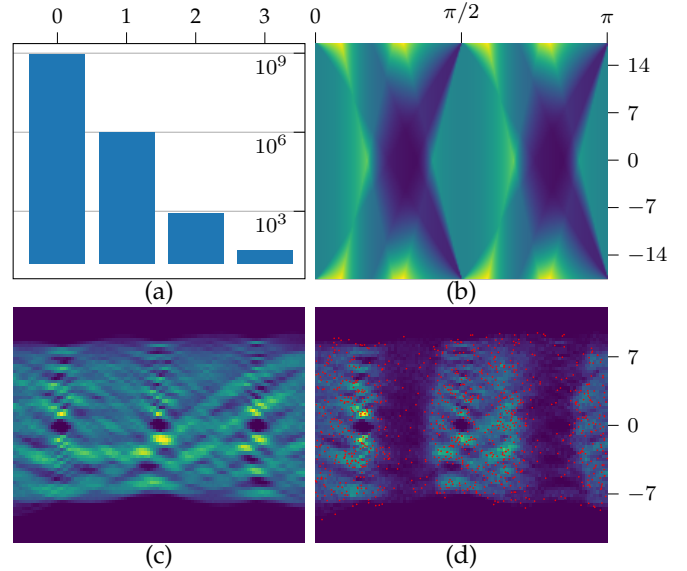}
           \vspace{-22pt}
        \caption{High-resolution PET. \textbf{a)} Number of lines of response in the sinogram of the scanner that detected $\{0,1,2,3\}$ pairs of photons. \textbf{b)} 
        Sensitivity $\xi$ of each line of response in the sinogram domain. Colorbar: \unit{1.4} to \unit{10}{\%} (dark to bright). \textbf{c)} Ground-truth sinogram. \textbf{d)} Data as acquired by the scanner (red samples) and histogram thereof \cite{iacobucci_monolithic_2021}.}
            
            \label{fig:pet}
        \end{figure}

\section{Proposed Framework
} \label{sec:proposed}


We now formulate our sensitivity-aware DE problem and build the method on a basis of B-splines on a grid. We approach bandwidth selection via proximable~regularization.

\subsection{Problem Formulation}\label{sec:problem}
    
        Let $\dens: \sinodomain \subset \reals^d \to \reals_{\geq 0}$, $d \in \naturals_{>0}$, be the pdf that corresponds to a phenomenon of interest. 
        Our aim is to estimate $\dens$ based on observations provided by physical sensors with a sensitivity that depends on their location within the compact domain $\sinodomain$.
        
        The sensitivity map $\sensi: \sinodomain \to \reals_{\geq 0}$ models the probability of the detection system to register an instance $\point \in \sinodomain$ of the phenomenon described by $\density$ (Figure~\ref{fig:pet}b). We assume that this phenomenon and the detection thereof are independent processes.
        The sensitivity map can either be measured experimentally or derived theoretically from some underlying detection model. Our requirements are that $\sensi$ is measurable and positive almost everywhere in $\sinodomain$. 
        Accordingly, we write the 
        density $\nu: \sinodomain \to \reals_{\geq 0}$
        of the detected events as
        \begin{equation}\label{eq:observed_density_norm}
        \obs \define \frac{\density \sensi }{ \int_{\sinodomain} \density \sensi}.
        \end{equation}
        This expression is equivalent to
        \begin{equation}\label{eq:observed_density}
        \obs = \frac{\densityu \sensi }{ \int_{\sinodomain} \densityu \sensi}
        \end{equation}
        for any $\densityu \propto \density$. For convenience, here we use $\densityu$ as an unnormalized proxy for the target density $\density$. 
        
        Note that the choosing of $\sensi$ is equivalent to the choosing of a measure; for example, of a reference measure for the vector exponential family that we formulate in \eqref{eq:original_density}. 
        Its presence in the normalization avoids the pointwise division typical of most weighted-density estimation methods analogous to \eqref{eq:kde}. This is because $\sensi$ does not need to be inverted, and because its effect is integrated over the support of the basis functions that we introduce next in Section \ref{sec:spline_param}. To emphasize this perspective, we adopt the notation
        \begin{equation}
        \Sensi(\bullet) = \int_\sinodomain \bullet \, \sensi(\point) \, \mathrm{d}\point.
        \end{equation}
        It can be interpreted as an expectation with respect to the Lebesgue–Stieltjes measure $\mu_{\sensiInt}$ associated to the cumulative function $\sensiInt$, where $\sensi$ is the derivative of $\sensiInt$ and, thus, the corresponding pdf, with $\sensi(\point) \mathrm{d}\point = \mathrm{d} \sensiInt(\point)=\mu_{\sensiInt}(\mathrm{d} \point)$.
        
        Our experimental observations are collected in a finite set $\Samples \subset \sinodomain$ of $\text{card}(\Samples)=\SamplesCard<\infty$ independent identically distributed (i.i.d.) realisations of $\obs$. Our goal is to recover $\density$ therefrom.

    \subsection{Spline Parameterization of the Density}\label{sec:spline_param}
        
        We parameterize our estimator $\densitye$ of $\densityu$ through coefficients $\coeffs \in \ell_2(\mathcal{M})$ by expressing it in terms of an exponential family of multidimensional splines on a finite, uniform grid $\mathcal{M} \subset \integers^d \cap \sinodomain$ with step size $\vmu \in \reals^d$. Specifically, we set
        \begin{equation}\label{eq:original_density}
            \densitye(\point; \coeffs) \define \exp{\left( \sum_{\m \in \mathcal{M}} \coeffsm \vbeta_\m (\point) \right)},
        \end{equation}
        where
        \begin{equation}
        \vbeta_{\index} (\point) \define \vbeta ( \point \oslash \vmu -   \index  ) \mbox{ for }\point \in \sinodomain .
        \end{equation}
        Whenever $\point \not\in \sinodomain$, $\vbeta_{\index}(\point)$ is specified by the BCs of the domain $\sinodomain$\tc{, see \eqref{eq:periodic_bcs} for a periodic example}. The symbols $\oslash$ and $\odot$ stand for element-wise division and multiplication, respectively. \tc{Note that we omit the dependency of $\vbeta_{\index}$  on $\vmu$ for conciseness. One decides on a $\vmu$ to choose the fineness of the grid with respect to the domain $\sinodomain$. The $\vmu$ acts by dilating the B-spline basis anisotropically and, thereby, the function as well. The type of BC is preserved upon such a dilation. }
        
        The tensor-spline basis $\vbeta: \reals^d \to \reals_{\geq 0}$ is built out of the one-dimensional B-splines $\beta^n: \reals \to \reals_{\geq 0}$ of degree $n$ in a separable manner, as
        \begin{equation}
        \vbeta(\point) \define \prod_{k=1}^d\beta^n(x_k).
        \end{equation}
        The B-spline of degree $n$ has a support of size $n+1$ and the approximation order $n+1$. 
        Closed-form expressions for several degrees can be found in \cite{unser_fast_1991}. In what follows, we omit the degree to simplify notation.

        \tc{By shifting the basis $\vbeta$ by $\m$ over $\mathcal{M}$, the B-spline expansion inside the exponential of \eqref{eq:original_density} can express any spline of the same degree $n$ and with the same (uniform) knots as per the theory in~\cite{schoenberg_i_j_contribution_1946}. 
        For example, piecewise-linear splines with knots at the integers are uniquely characterized by $\sum_{\m \in \mathcal{M}} \coeffsm \vbeta_\m (\point)$ with $\vmu=\mathbf{1}$, and $\vbeta$ stemming from the triangle function $\beta^{1}(x)=\max(1-|x|,0)$. For arbitrary $\vmu$, the knots are at $\vmu \mathbb{Z}$ and the resulting function is an anisotropic dilation of a spline, which is also a spline.}

        \tc{While standard logsplines are defined on the entire real line with nonuniform knots, expression \eqref{eq:original_density} is defined on multidimensional domains with general BCs, but on a uniform grid. One advantage is that }B-spline expansions are numerically efficient, owing to their favorable tradeoff between support and accuracy \cite{unser_splines_1999}. Many relevant operations 
        can be written as convolutions under these expansions \cite{unser_fast_1991}. They are also well equipped to handle finite domains subject to appropriate BCs. 
        \tc{One incorporates these by performing the convolution-based operations on the coefficients under said BCs, which is especially convenient for periodic domains because they can be computed using the FFT. } One consequence \tc{of this efficiency} is that the density can be quickly evaluated without loss of accuracy. For example, evaluations on uniform grids can be computed exactly via convolutions. This is desirable for iterative methods in imaging, where repeated evaluations might be required. The exponential guarantees that \eqref{eq:original_density} is nonnegative. \tc{In Section~\ref{sec:nuclear_norm}, we further comment on how our (regularized) DE method relates to standard logsplines and to the concept of free knots.}

        To illustrate the advantages of using splines on a grid, let us define a sampling operator $\samp_{\mathcal{M}_s}$ that takes in a function $h: \sinodomain \to \reals^q$ and evaluates it on a grid $\mathcal{M}_s \subset \sinodomain$. 
        This grid is upscaled from the original $\mathcal{M}=\mathcal{M}_1$ to a finer scale $s$. One writes that
        \begin{equation}
        [\samp_{\mathcal{M}_s}\{ h\}]_\m = h\left(\frac{\vmu}{s} \odot \m\right).
        \end{equation}
        The codomain of $\samp_{\mathcal{M}_s}$ is thus  $\reals^{|\mathcal{M}_s| \times q}$.
        The sampling operator allows us to write the evaluation of the density estimate as the convolution
        \begin{equation}\label{eq:grid}
        \samp_{\mathcal{M}_s}\{ \log \densitye (c) \} = \ker_s^{(\mathbf{0})} * \coeffs_{\uparrow s }
        \end{equation}
        in $d$ dimensions with the BCs corresponding to $\sinodomain$. The vertical arrow subscript in $\coeffs_{\uparrow s }$ refers to the upsampling of $\coeffs$ by expansion with $(s-1)$ zeros, so that $\coeffs_{\uparrow s }[\m']=\coeffs[\m]$ if $\m'=s\m$ and $\coeffs_{\uparrow s }[\m']=0$ otherwise.
        The discrete convolution kernel $\ker_s^{(\mathbf{0})} $ in \eqref{eq:grid} is separable, with
        \begin{equation}\label{eq:tensor_spline}
        \ker_s^{(\mathbf{n})} [\m] = 
        \prod_{k=1}^{d}
        \, \frac{\partial^{n_k} \beta}{\partial \point_k^{n_k}} (\point) \bigg\rvert_{\point=\m/s},
        \end{equation}
        which is the B-spline filter that corresponds to the evaluation of the derivative of order $\mathbf{n}=(n_k)$, $n_k \in \naturals_{\geq 0}$.
        This notation will be useful for the Hessian-based regularizer because the evaluation of derivatives can also be written in terms of convolutions. 

    \subsection{Likelihood as Data Fidelity}
Aiming at the recovery of $\density$ from the measurements of $\obs$, we formulate an optimization problem with respect to the coefficients 
$\coeffs \in \mathcal{C}=\ell_2(\mathcal{M})$.


The data fidelity is based on the likelihood of observing the i.i.d. samples $\sample \in \Samples$ and is given by
\begin{equation}
L_{\SamplesSub } (\coeffs) \define \frac{1}{ \Sensi\left( \densitye(c) \right)^{\SamplesCard}} \prod_\qinq \densitye (\sample; \coeffs) \sensi  (\sample) \approx  \prod_\qinq \obs(\sample; \coeffs).
\end{equation}
In particular, we work with the corresponding log-likelihood
\begin{align}\label{eq:loglikelihood}
\log(L_{\SamplesSub })(c) = & 
\sum_{\m \in \mathcal{M}} \sum_\qinq \coeffsm \vbeta_\m(\sample)
- \SamplesCard \log \left( \Sensi\left( \densitye(\cdot; c) \right)\right) \nonumber \\
&+ \sum_\qinq  \log\sensi (\sample).
\end{align}
For tidiness, we shall often omit the dependency of the density on $c$, simply writing $\densitye$ instead of $\densitye(\cdot; c)$.

The log-likelihood has derivatives
\begin{align}\label{eq:loglikelihood_derivative}
\mathbf{g}_\k(c)
\define & \partial_{\coeffs[\indexk]} \log(L_{\SamplesSub }) \nonumber \\
= & \sum_\qinq \vbeta_\indexk(\sample)-\frac{\SamplesCard}{\Sensi(\densitye)} \Sensi \left(\vbeta_{\indexk} \densitye\right)
\end{align}
with respect to every $\coeffs[\indexk]$, $\k \in \mathcal{M}$. It also has the  Hessian $\mathrm{\mathbf{H}}(c)$ of size $|\mathcal{M}|^{2}$ with elements
\begin{align}\label{eq:likelihood_hessian}
\left[\mathrm{\mathbf{H}}\right]_{\indexk, \indexl}(c) \define & \partial^2_{\coeffs[\indexk], \coeffs[\indexl]} \log(L_{\SamplesSub}) \nonumber \\
 = & -\frac{\SamplesCard}{\Sensi(\densitye)}   \bigg( \Sensi ( \vbeta_{\indexk} \vbeta_{\indexl} \densitye) \nonumber \\
 & -  \frac{\Sensi \left(\vbeta_{\indexk} \densitye\right) \Sensi \left(\vbeta_{\indexl} \densitye\right)}{\Sensi(\densitye)}  \bigg).
\end{align}
The Hessian of the log-likelihood is negative-definite because
\begin{align}
\sum_{\indexk \in \mathcal{M}} & \sum_{\indexl \in \mathcal{M}}  \test[\indexk] \test[\indexl] \mathrm{H}_{\indexk, \indexl}(c) = 
\nonumber \\  &-\frac{1}{\Sensi(\densitye)}  \Sensi \bigg( \bigg(\sum_{\m \in \mathcal{M}} \testm \vbeta_\m \nonumber  
\nonumber \\  &- \Sensi \bigg( \sum_{\m \in \mathcal{M}} \testm \vbeta_\m \densitye \bigg) \bigg)^2 \densitye \bigg) 
\nonumber  \\
& < 0
\end{align}
for all $\test \neq 0$ since $\densitye, \sensi \geq 0$ and the innermost term cannot be zero because the $\vbeta_\m$ are linearly independent. The log-likelihood is therefore strictly concave. Hence, \eqref{eq:loglikelihood} has a unique maximum at which the coefficients parameterizing $\densitye$ are said to be optimal. Here, existence of the maximum-likelihood estimate is guaranteed for the non-degenerate cases (when there are enough data and a constant solution is not better).

The normalization factor $\Sensi(\densityu)=\int_{\sinodomain} \densityu \sensi$ in \eqref{eq:observed_density} constrains the resulting density in two ways. The first is through the use of the B-spline basis because the effect of a single sample $\sample$ on a single $\coeffs[\index]$ is spread throughout the support of the basis function. 
The second is through the \textit{a priori} sensitivity, which implicitly weights the density according to the probability of detection. As the choosing of $\sensi$ is equivalent to the choosing of a measure, the approach retains many of the interesting properties of the original 1D logsplines \cite{stone_large-sample_1990}. 

Notice that $\sensi$ plays its role \tc{in the optimization} exclusively through the normalization found in \eqref{eq:observed_density} or \eqref{eq:loglikelihood}-\eqref{eq:loglikelihood_derivative} because the derivatives would be independent of $\sensi$ if the observed $\obs$ went unnormalized. \tc{The gradient is driven by the local data points and by the local (over the basis support) contribution to the normalization integral relative to the sensitivity. 
The framework accepts (without change) any sensitivity function that is nonnegative and for which the integrals in \eqref{eq:loglikelihood_derivative} can be computed; it does not require a (multiplicative) inverse (cf.~\eqref{eq:kde}). Some computations can be sped up (or performed more accurately) when the sensitivity is smooth or, even better, if it is expressed in the same form as \eqref{eq:original_density}. 
} 







\begin{figure}
    \centering
    \input{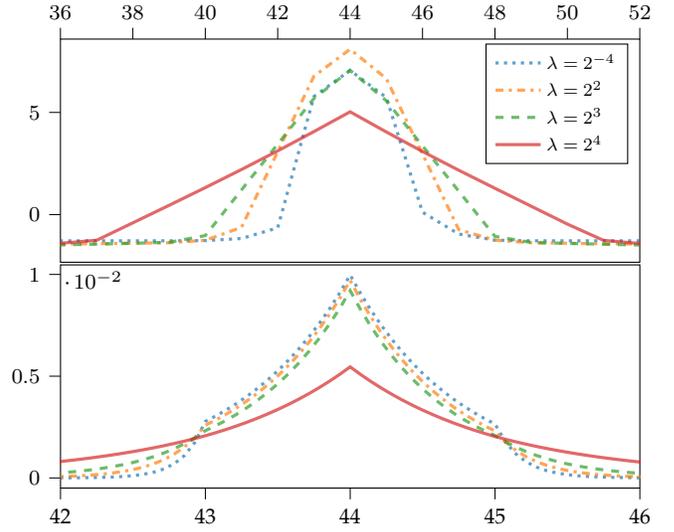}
    \vspace{-12pt}
    \caption{Effect of the regularization on the estimate for splines of degree $n=1$. A slice of the recovered density along the first axis ($d=2$) is shown. Top: logarithm of the density (spline). Higher $\lambda$ results in "less" knots. Bottom: density (zoomed in because it is the exponential of the spline).}
    \label{fig:sparsity}
    \end{figure}
    
\subsection{Proximal-Based Regularization of the Underlying Spline}
\label{sec:regularization}
To further constrain the problem, we add a non-differentiable regularization term $R$ to the log-likelihood \eqref{eq:loglikelihood} of the data. This results in the optimization problem:
\begin{equation}\label{eq:min_problem}
    \opticoeffs = \arg\min_{\coeffs\in\mathcal{C}} \func(c)
\end{equation}
\begin{equation}
\func(\coeffs) \define \lambda R(c) - \log(L_{\SamplesSub})(c),
\end{equation}
where $\lambda \in \reals_{> 0}$ controls the balance between data fidelity (likelihood) and prior knowledge (regularizer). We enforce the desired \textit{a priori} behavior directly on the spline formulation as $R(\log(\densitye(c)))$ because it has been found experimentally that regularizing the logarithm of the density is often a good approach to capture the multimodality of data \cite{koenker_density_2007, kooperberg_study_1991}.

\subsection{Nuclear Norm of the Hessian for Knot Sparsity}
\label{sec:nuclear_norm}
\def\mata{\mathrm{\mathbf{A}}}

 Our choice of regularization is based on the Schatten $p$-norm of the Hessian ($\mathcal{H}$) of the log-density over the domain, whereby
\begin{equation}\label{eq:reg_term}
R(\coeffs) \define \int_{\sinodomain} \norm{\mathcal{H}\{\log(\densitye)\}}_{S_p}, 
\end{equation}
where
\begin{equation}\label{eq:schatten_norm}
\norm{\mata}_{S_p} := \norm{\sigma(\mata)}_p
\end{equation}
is the Schatten $p$-norm of a matrix $\mata$ with a vector $\sigma \left( \mata \right)$ of singular values, and $p \in [1, \infty]$.

We develop theory and software for general $p$. In practice, however, our choice will be the nuclear norm, which corresponds to $p=1$. It is also known as the trace norm $\norm{\mata}_{S_1}:=\tr\left( \sqrt{\mata^{^{\mathsf{H}}}\mata} \right)$ in general, 
and
$\norm{\mata}_{S_1}=\tr\left( \mata \right)$ for square, positive-semidefinite matrices. Since the nuclear norm is a convex envelope of the rank function, $p=1$ serves here as a convex surrogate for the promotion of low-rank Hessians with one main singular value---or principal curvature. The nuclear norm is also invariant to isometries because it is a norm of the vector of singular values.
    
   Our choice encourages the affineness of the splines whenever information is lacking, which translates into an exponential behavior of the density. This happens because sparse Hessians are promoted. The resulting behavior tackles two challenges of \tc{standard} logsplines at once.

   \subsubsection{Tail Behavior} The regularization encourages a linear behavior also at the tails of the density where few samples, if any, are available. This property is known to reduce the variance at the borders \cite{mcdonald_review_2021}. In classical logspline fitting, it has to be specifically enforced by mixing splines of several degrees \cite{kooperberg_comparison_2004}.

    \subsubsection{Bandwidth Adaptability} Playing a similar role to adaptiveness in KDE, knot placement and knot deletion are important steps of typical logspline fitting that involve the optimization of certain information criteria \cite{kooperberg_comparison_2004, truong_chapter_2005}. For $d=1$, our approach automatically \tc{deactivates}/eliminates the least-relevant knots when the B-splines are linear---it induces knot sparsity. For higher $d$ and arbitrary degree, it promotes the sparsity of the Hessian and the linearity of the logarithm 
    when the data are scarce (Figure~\ref{fig:sparsity}). In turn, this sparsity reduces the influence of the coarseness of the grid (Figure~\ref{fig:lambda_vs_mse}). \tc{See Appendix~A, for a more theoretical description of the effect of the regularization and its relation with free knots.} \tc{Note that the convolution-based computations facilitate starting with fine grids of knots.}

In practice, we approximate $R$ in a discrete setting by evaluating the Hessian on the grid with the mixed $S_p$-$\ell_1$ norm
\begin{align}\label{eq:schatten_grid}
R(\coeffs) \approx & 
\norm{\samp_{\mathcal{M}_1} \{ \mathcal{H}\{ \log \densitye (c) \} \} }_{S_p, 1} \nonumber \\
& =  \sum_{\m \in \mathcal{M}} \norm{\mathcal{H}\{ \log \densitye (c) \}(\m) }_{S_p},
\end{align}
where $\samp_{\mathcal{M}_1} \{ \mathcal{H}\{ \log \densitye (c) \} \} : \mathcal{C} \to \reals^{|\mathcal{M}_1| \times d^2}$. In Section \ref{sec:opti}, we show how to evaluate the last term of \eqref{eq:schatten_grid} efficiently, as well as its adjoint and proximal operator.

Beware that the Hessian of this section should not be confused with the Hessian of the log-likelihood.

\section{Optimization of the Proposed Framework}\label{sec:opti}
The functional $\func$ inherits strict convexity from $\left(-\log(L_{\SamplesSub}) \right)$ because $R$ is convex. We thus approach the problem in \eqref{eq:min_problem} with an accelerated proximal-gradient algorithm in mind \cite{liang_improving_2022,lee_proximal_2014}.
The gradient of $\log(L_{\SamplesSub})$ is computed from \eqref{eq:loglikelihood_derivative}. The proximal operator of the regularizer $R$ in \eqref{eq:schatten_grid} can also be computed efficiently, as detailed in Section \ref{sec:prox}.

Although the negative log-likelihood $\left(-\log(L_{\SamplesSub})\right)$ is twice continuously differentiable, 
it is only locally gradient-Lipschitz and locally strongly convex. 
In effect, unless the underlying domain is compact, the norm of its second derivatives cannot be upper- or (positively) lower-bounded because we are working with an exponential family.
Even for a compact domain, the worst-case bounds will be too loose elsewhere. The consequence is that it is inefficient to set globally the size of the descending step of the proximal-gradient algorithm. It needs to be adapted at each step (see our bound for the local Lipschitz constant in Figure~\ref{fig:lipschitz}).

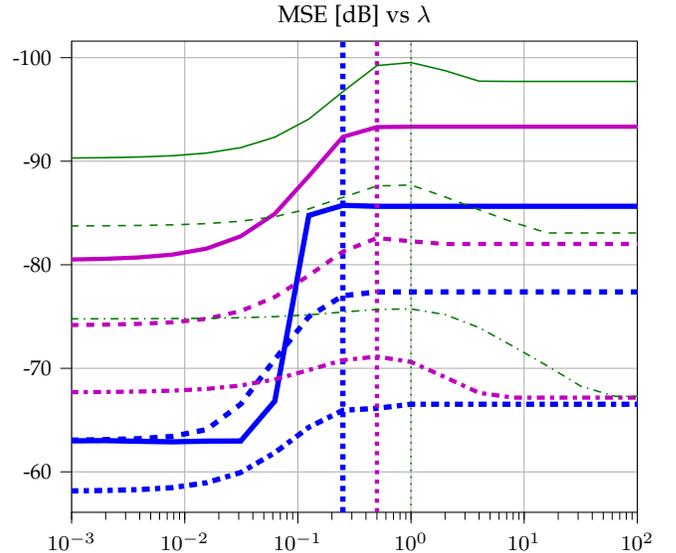
\begin{figure}
\centering
\begin{tikzpicture}

\definecolor{darkgray176}{RGB}{176,176,176}
\definecolor{darkviolet1910191}{RGB}{191,0,191}
\definecolor{green01270}{RGB}{0,127,0}
\def\ys{-4pt}
\begin{axis}[
title style={yshift=\ys,},
log basis x={10},
tick align=outside,
tick pos=left,
x grid style={darkgray176},
xmin=0.001, xmax=100,
xmode=log,
xtick style={color=black},
y grid style={darkgray176},
ymin=56.0928609072292, ymax=95., ymax=101.596824451834,
ytick style={color=black},
ytick = {60,70,80,90,100},
yticklabels = {-60,-70,-80,-90,-100},
title={\small MSE [dB] vs $\lambda$},
width=0.5\textwidth,
grid=major
]
\addplot [line width=2., blue, dash dot]
table {%
0.0009765625 58.1612228865294
0.001953125 58.1994995011506
0.00390625 58.2821500804422
0.0078125 58.4771118582837
0.015625 58.9515897019732
0.03125 59.941652678828
0.0625 61.8457660107869
0.125 64.3074790572092
0.25 65.9450283265136
0.5 66.1369328329752
1 66.5303905657543
2 66.527878564045
4 66.5275886743416
8 66.5275886743416
16 66.5275886743416
32 66.5275886743416
64 66.5275886743416
128 66.5275886743416
};
\addplot [line width=2., blue, dashed]
table {%
0.0009765625 63.0648574704753
0.001953125 63.1107344231109
0.00390625 63.205968159412
0.0078125 63.4243717358554
0.015625 64.0861794675726
0.03125 66.5400374689138
0.0625 70.8550641011187
0.125 74.9908908125449
0.25 77.0122656541661
0.5 77.3742120745549
1 77.3782092601069
2 77.3785597482059
4 77.3791550400462
8 77.3796304241561
16 77.3797460080531
32 77.3797460080531
64 77.3797460080531
128 77.3797460080531
};
\addplot [line width=2., blue]
table {%
0.0009765625 62.99239597873
0.001953125 63.0145894773027
0.00390625 62.95297475627
0.0078125 62.8987133555727
0.015625 62.9684037685684
0.03125 62.978144895351
0.0625 66.8340707387674
0.125 84.7638372346591
0.25 85.7361305091763
0.5 85.6428609599361
1 85.6449813072937
2 85.6452658530185
4 85.6446614992025
8 85.6450399947418
16 85.6454751004848
32 85.6460544228138
64 85.6468762118109
128 85.6468762118112
};
\addplot [line width=2., blue, dotted]
table {%
0.25 56.0928609072292
0.25 101.596824451834
};
\addplot [ultra thick, darkviolet1910191, dash dot]
table {%
0.0009765625 67.6930826978731
0.001953125 67.7141626522949
0.00390625 67.7561140618941
0.0078125 67.839916235963
0.015625 68.0087195326995
0.03125 68.3366489050011
0.0625 68.928138005761
0.125 69.828811957362
0.25 70.7875073997322
0.5 71.1294123008078
1 70.6154666423882
2 69.1436221934888
4 67.6254844743067
8 67.1682251802917
16 67.1681982984772
32 67.1681973437382
64 67.1681973437382
128 67.1681973437382
};
\addplot [ultra thick, darkviolet1910191, dashed]
table {%
0.0009765625 74.1780958250757
0.001953125 74.2149340207548
0.00390625 74.289972574291
0.0078125 74.4459312548287
0.015625 74.7819708756373
0.03125 75.5246075225372
0.0625 76.8888497195499
0.125 78.9649934902553
0.25 81.2747511427994
0.5 82.5784821463325
1 82.2674666154489
2 81.9995419482057
4 82.010840984176
8 82.0108241376141
16 82.0108240196327
32 82.0108240196327
64 82.0108240196327
128 82.0108240196327
};
\addplot [ultra thick, darkviolet1910191]
table {%
0.0009765625 80.5076341362304
0.001953125 80.5731943244168
0.00390625 80.7068500485496
0.0078125 80.983390662616
0.015625 81.5645736886763
0.03125 82.7509332499406
0.0625 84.9197726228108
0.125 88.5640058848997
0.25 92.3519893825289
0.5 93.2947530222363
1 93.3345917520954
2 93.3347182701398
4 93.3348746632249
8 93.3349781060979
16 93.3350368221942
32 93.3350414236087
64 93.3350414236087
128 93.3350414236087
};
\addplot [ultra thick, darkviolet1910191, dotted]
table {%
0.5 56.0928609072292
0.5 101.596824451834
};
\addplot [semithick, green01270, dash dot]
table {%
0.0009765625 74.7784130346756
0.001953125 74.7820888196765
0.00390625 74.7893900364927
0.0078125 74.8038446802904
0.015625 74.8321678608932
0.03125 74.8866629347101
0.0625 74.9878047682278
0.125 75.1642424640296
0.25 75.4323856391814
0.5 75.6871654261229
1 75.7332956301854
2 75.1924276061778
4 73.9325255349226
8 72.12163881566
16 70.2380108008956
32 68.2493702896944
64 67.2904990200643
128 67.2904912876089
};
\addplot [semithick, green01270, dashed]
table {%
0.0009765625 83.7487505603633
0.001953125 83.7640570184057
0.00390625 83.7945536115802
0.0078125 83.8550540071581
0.015625 83.9745496753785
0.03125 84.2062162194884
0.0625 84.6430549340273
0.125 85.3955506897103
0.25 86.5083220520682
0.5 87.6082281592263
1 87.7033995375117
2 86.5529549628851
4 85.3027157002209
8 84.1051818654062
16 83.069991005737
32 83.0700025161156
64 83.0700025161156
128 83.0700025161156
};
\addplot [semithick, green01270]
table {%
0.0009765625 90.3101527469338
0.001953125 90.3413951868363
0.00390625 90.4040467422218
0.0078125 90.5300590335807
0.015625 90.7859800920415
0.03125 91.3099643842571
0.0625 92.3218434639471
0.125 94.0701162707898
0.25 96.7155624554543
0.5 99.2363393303731
1 99.5284624725342
2 98.7480719138747
4 97.7296677984705
8 97.7060167480605
16 97.7060257114949
32 97.7060257114948
64 97.7060257114949
128 97.7060257114949
};
\addplot [semithick, green01270, dotted]
table {%
1 56.0928609072292
1 101.596824451834
};
\end{axis}

\end{tikzpicture}

    \vspace{-10pt}
    \caption{Mean-squared error of the regularized-density splines estimate of (Uniform + Gaussian + Laplacian) compound density as a function of the regularization parameter $\lambda$ for three numbers of samples $\{10^k \, | \, k \in \{2,3,4\} \}$ (blue fat, magenta medium, green thin) combined with three grid sizes $\{(44k,44k)  \,| \, k \in \{1,2,3\} \}$ (mixed, dashed, solid). Vertical lines at $\{0.25,0.5,1\}$ mark the $\lambda$ with minimal error for each grid size.
    }
    \label{fig:lambda_vs_mse}
\end{figure}

\subsection{Derivation of an Adaptive Lipschitz Stepsize}

To take advantage of the convergence rates of accelerated algorithms, we devised an adaptive strategy based on Lipschitz bounds. The appeal of the approach is based on the observation that the tensor
\def\g{\mathbf{g}}
\def\nonoutm{\mathrm{\mathbf{D}}}
\def\outm{\mathrm{\mathbf{F}}}
\def\outv{\mathbf{f}}
\begin{equation}
 \left[\outm\right]_{\k, \indexl} (c) = \frac{\Sensi(\vbeta_{\indexk}\densitye)\Sensi(\vbeta_{\indexl}\densitye)}{\Sensi(\densitye)^2},
\end{equation}
$\k,\indexl \in \mathcal{M}$, in the Hessian \eqref{eq:likelihood_hessian} 
can be written as the outer product $ \outm = \outv \otimes \outv
$, where
\begin{equation}
 \left[\outv\right]_{\k}(c) = \frac{\Sensi(\vbeta_{\indexk}\densitye)}{\Sensi(\densitye)}
\end{equation}
is a value that turns out to be an essential ingredient of the gradient \eqref{eq:loglikelihood_derivative}. The remaining term 
\begin{equation}
\nonoutm = \hess - \outm,
\end{equation}
originating from $\Sensi(\vbeta_{\indexk} \vbeta_{\indexl} \densitye)\Sensi(\densitye)^{-1}$ in \eqref{eq:likelihood_hessian} may have multiple nonzero singular values in general, but is always banded, symmetric and positive-definite.

We formulate the adaptive strategy by bounding the best Lipschitz constant $\lipsbest$ of the gradient $\mathbf{g}$ at $\coeffs$ with the Hessian as
\begin{equation}\label{eq:lips_ineq}
\lipsbest \leq \norm{\hess}_2 \leq \norm{\nonoutm}_2 + \norm{\outm}_2.
\end{equation}
The rightmost term is computed by exploiting the outer product, leading to
\begin{equation}
\norm{\outm}_2 = \norm{\outm}_\mathrm{F} = \norm{\outv}_2^2,
\end{equation}
where the subscripts ``$\mathrm{F}$'' and ``$2$'' denote the Frobenius and spectral norms, respectively. 
To bound $\norm{\nonoutm}_2$, we use Gershgorin's circle theorem because we know $\nonoutm$ will usually be diagonally dominant. This yields a bound $D_2(c)$ defined as
\begin{equation}\label{eq:gersh_circle}
\norm{\nonoutm}_2 \leq \max_\k \Big( \left[\nonoutm\right]_{\k,\k} + \sum_{\indexl \neq \k} |\left[\nonoutm\right]_{\k,\indexl}| \Big) = D_2.
\end{equation}
Notice that there are at most $|\mathcal{M}|(2n+1)^d$ nonzero elements, and only $|\mathcal{M}|(2n+1)^d/2+|\mathcal{M}|/2$ of them need to be computed because of the symmetry of $\nonoutm$. The big number of zero elements is due to the small support of the B-spline basis.
A more general bound is
\begin{equation}\label{eq:general_bound}
\norm{\nonoutm}_2 \leq \norm{\nonoutm}_{\mathrm{F}} \leq \sqrt{|\mathcal{M}|(2n+1)^d}\left( \int_\sinodomain \beta \beta \right)^{2d}\bar{\obs}
\end{equation}
with $\bar{\obs}=|\mathcal{M}|^{-1} \sum_{\m \in \mathcal{M}} \max(w_{\m}(\nu)) \leq \max_{\point \in \sinodomain}(\obs(\point))$, where $w_{\m}$ is a mask of size $(2n+1)^d$ around $\m$. We resort to \eqref{eq:general_bound} in the rare cases where \eqref{eq:gersh_circle} is a looser bound.

To set the stepsize of the proximal-gradient descent, we use the inverse of the bound $\lips(c)$ that results from combining the previous bounds as
\begin{equation}\label{eq:lips_bound}
\lips = \norm{\outm}_2 + D_2 \geq \lipsbest.
\end{equation}

\subsection{Filters for the Hessian-Schatten Norm}
\def\inx{\mathbf{n}}
To compute \eqref{eq:schatten_grid}, we harness the convenience of splines on a grid. Evaluating the Hessian of the logarithm of the density estimator at all grid points involves $d(d+1)/2$ convolutions. The $\inx$th component of the Hessian at grid point $\m \in \mathcal{M}$ can be expressed as
\begin{equation}\label{eq:schatten_grid_spline}
\left[ \mathcal{H}\{ \log \densitye (\coeffs) \}(\m) \right]_{\inx} = \left( \ker^{(\inx)}_{1} * \coeffs \right)[\m].
\end{equation}
The filters $\ker^{(\inx)}_{1}$ are as described in Section \ref{sec:spline_param}. We clarify that the entry $\inx=(n_1, n_2) \in \{1 \ldots d\}^2$ of the Hessian corresponds to the partial derivatives $\partial^2/\partial {x_{n_1}}\partial{x_{n_2}}$ of $\log \densitye$. A single convolution is therefore enough to compute the $\inx$th Hessian element at all grid points. 
They are tensor combinations of filters that are similar to centered finite-difference kernels such as $[1,\fbox{-2},1]$. A recurrent expression for one-dimensional B-spline filters can be found in \cite{unser_b-spline_1993}.

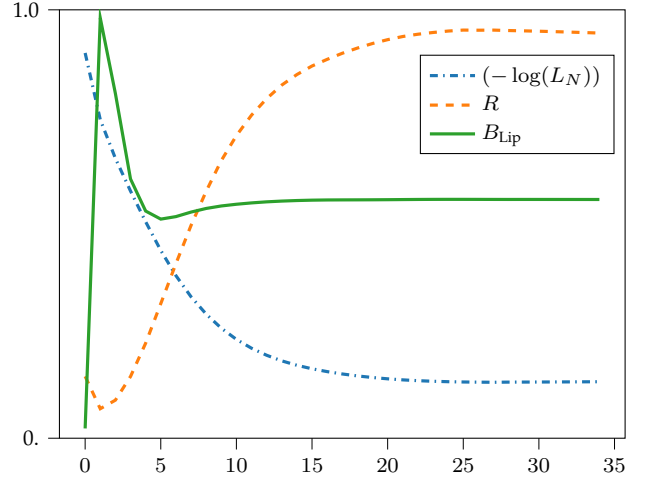
\begin{figure}
    \centering
\begin{tikzpicture}[every plot/.append style={very thick}]

\definecolor{darkgray176}{RGB}{176,176,176}
\definecolor{darkorange25512714}{RGB}{255,127,14}
\definecolor{forestgreen4416044}{RGB}{44,160,44}
\definecolor{steelblue31119180}{RGB}{31,119,180}

\begin{axis}[
width=.5\textwidth,
height=0.4\textwidth,
tick align=outside,
tick pos=left,
x grid style={darkgray176},
xmin=-1.7, xmax=35.7,
xtick style={color=black},
y grid style={darkgray176},
ymin=7950, ymax=8280,
ytick={7950, 8280},
yticklabels={0., 1.0},
ytick style={color=black}
]
\addplot [semithick, steelblue31119180, dash dot]
table {%
0 8246.78072103498
1 8195.41709535889
2 8165.58549793931
3 8140.4127936265
4 8116.76413949412
5 8094.66662443288
6 8075.29412842025
7 8058.98710055373
8 8045.63015147619
9 8034.77708466083
10 8026.13449217182
11 8019.3315164228
12 8013.91117437498
13 8009.58736302261
14 8006.18818288104
15 8003.48613270729
16 8001.31096523674
17 7999.49603713724
18 7998.00214515157
19 7996.77003462031
20 7995.75347394073
21 7994.94958830744
22 7994.33076309177
23 7993.86733656405
24 7993.53682061416
25 7993.31912297386
26 7993.18708282776
27 7993.09794003351
28 7993.22010217206
29 7993.25239824452
30 7993.29275172922
31 7993.3404914571
32 7993.39454345508
33 7993.45349123725
34 7993.51640643111
};
\label{plot_one}
\end{axis}
\begin{axis}[
width=.5\textwidth,
height=0.4\textwidth,
tick align=outside,
tick pos=right,
axis y line*=right,
axis x line=none,
x grid style={darkgray176},
xmin=-1.7, xmax=35.7,
xtick style={color=black},
y grid style={darkgray176},
ymin=175, ymax=240,
ytick style={color=black},
hide y axis,
]
\addplot [semithick, darkorange25512714, dashed]
table {%
0 184.383494060346
1 179.476016774458
2 180.774327473085
3 184.365117931345
4 189.421127671264
5 195.41714618645
6 201.47866258481
7 207.281204193835
8 212.461107202571
9 217.121968124901
10 220.915136174593
11 224.02297438621
12 226.524665626309
13 228.586911542414
14 230.181807851034
15 231.478507707045
16 232.485912262977
17 233.409138860615
18 234.198356750622
19 234.893219087747
20 235.475871988626
21 235.95389085635
22 236.300615223907
23 236.564871030174
24 236.782634747638
25 236.930128849207
26 236.909811579547
27 236.925908786596
28 236.85971986174
29 236.811089611755
30 236.756131245596
31 236.694461082076
32 236.625486864299
33 236.55320216606
34 236.480820897759
};
\label{plot_two}
\end{axis}
\begin{axis}[
width=.5\textwidth,
height=0.4\textwidth,
tick align=outside,
tick pos=left,
axis y line*=right,
axis x line=none,
x grid style={darkgray176},
xmin=-1.7, xmax=35.7,
xtick style={color=black},
y grid style={darkgray176},
ymin=5, ymax=57.5,
ytick style={color=black},
every y axis line/.style={xshift=-1.5cm}, 
every tick/.style={xshift=-1.5cm}, 
every y tick label/.style={xshift=-1.5cm},
every outer y axis line/.style={xshift=-1.5cm},
legend style={at={(axis cs:35,52)},anchor=north east},
legend cell align={left},
hide y axis,
]
\addlegendimage{/pgfplots/refstyle=plot_one, steelblue31119180, dash dot, very thick}\addlegendentry{$ \left(- \log(L_{\SamplesSub})\right)$}
\addlegendimage{/pgfplots/refstyle=plot_two, darkorange25512714, dashed, very thick}\addlegendentry{$R$}
\addplot [semithick, forestgreen4416044, very thick]
table {%
0 6.19834710743802
1 56.4930599331212
2 47.2441982141366
3 36.7706275189887
4 32.8418135422059
5 31.851577263938
6 32.158265792793
7 32.7125918320781
8 33.1572641032697
9 33.4625003704975
10 33.672844561668
11 33.8382954611284
12 33.9686198407854
13 34.061162640817
14 34.128808715486
15 34.1738930769534
16 34.1989843123185
17 34.2077235142121
18 34.2131339157121
19 34.2203935022107
20 34.2315537486404
21 34.2471749212596
22 34.2611184162142
23 34.2699340657183
24 34.2731194163243
25 34.2723366643483
26 34.2688613507414
27 34.2639835318591
28 34.2610007241227
29 34.2630004806088
30 34.2619465020032
31 34.2607315759508
32 34.2593619985567
33 34.257846609534
34 34.2561762859187
};
\addlegendentry{$\lips$}\end{axis}

\end{tikzpicture}
    \vspace{-10pt}
    \caption{Evolution until convergence of the data and regularization terms, and of the bound $\lips (c) \geq \lipsbest$ of the local Lipschitz constant as a function of the number of iterations ($y$ scales are normalized).} 
    \label{fig:lipschitz}
\end{figure}

\begin{figure}[ht]
    \centering
    \input{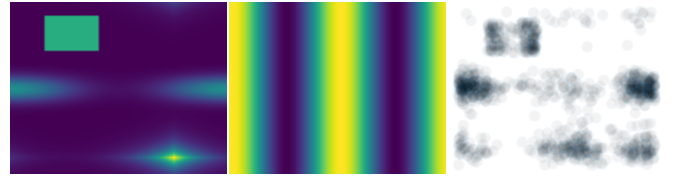}
    \vspace{-8pt}
    \caption{Left to right: The UGL distribution, the sS sensitivity, and $10^3$ samples out of their combination. }
    
    \label{fig:samples}
\end{figure}

\subsection{Proximal of the Hessian-Schatten Norm}\label{sec:prox}
We compute the proximal operator of \eqref{eq:schatten_grid}, \eqref{eq:schatten_grid_spline} by solving the corresponding minimization problem, which yields
\begin{eqnarray}\label{eq:prox_hessian_schat}
\prox_{\tau R}(\tilde{\coeffs}) &=& \arg \min_{\coeffs \in\mathcal{C}} \Big(  \, \frac{1}{2}\norm{\coeffs-\tilde{\coeffs}}_2^2 \nonumber\\ 
&&\mbox{} + \tau \norm{\samp_{\mathcal{M}_1} \{ \mathcal{H}\{ \log \densitye (c) \} \} }_{S_p, 1} \Big). 
\end{eqnarray}
We do it iteratively, using a gradient-projection algorithm based on the maximization of the dual formulation of the problem in \eqref{eq:prox_hessian_schat}. We refer to \cite{beck_fast_2009} for a description of the approach. 
To implement it,
we require three \textit{ad hoc} ingredients: the adjoint of the operator $\samp_{\mathcal{M}_1} \{ \mathcal{H}\{ \log \densitye \} \}$ inside the $S_p$-$\ell_1$ norm, the projection onto the unit ball of the dual $S_q$-$\ell_\infty$ norm ($1/p+1/q=1$), and an adequate stepsize.

\subsubsection{Adjoint} We derive the adjoint of the evaluation $\samp_{\mathcal{M}_1} \{ \mathcal{H}\{ \log \densitye \} \}$ of the Hessian of the spline on the grid by enforcing the adjoint definition. It is the sum of the convolutions with the adjoint of the filters under the adjoint BCs, as given by
\begin{equation}
\left( \samp_{\mathcal{M}_1} \{ \mathcal{H}\{ \log \densitye \} \} \right)^*(\mata) 
=  \sum_{\mathbf{k} \in \{1 \twodots d\}^2}  \left({\ker^{(\indexk)}_{1} }\right)^* * \left[\mata\right]_{\cdot, \mathbf{k}},
\end{equation}
where $\mata \in \reals^{|\mathcal{M}_1| \times d^2}$, the superscript $^*$ refers to the adjoint, and the dot in subscript $[]_{\cdot, \indexk}$ indicates that the convolution is performed over the $\reals^{|\mathcal{M}_1|}$ dimension. In principle, the impulse response of the adjoint $\left({\ker^{(\indexk)}_{1} }\right)^*$ of a filter would have to be the reversed impulse response of the original filter along the appropriate axes, but in fact it is simply that of the original filter because of considerations on symmetry.

\subsubsection{Projection} The projection of a matrix onto the unit ball of the Schatten $q$-norm requires one to decompose the matrix into singular values and project them onto the unit ball of the $q$-norm \cite[Proposition~1 and Equation~(34)]{lefkimmiatis_hessian_2013}. The matrix is then recomposed using the original singular vectors. For small $d$, closed-form solutions exist, which leads to a fast implementation. 

\subsubsection{Stepsize} To maximize the objective function that is dual to that in \eqref{eq:prox_hessian_schat}, we need to choose the stepsize of the gradient-ascent algorithm. We set it according to an upper bound for the Lipschitz constant of the gradient of the dual. By generalizing the argument in \cite[Proposition~2]{lefkimmiatis_hessian_2013} to arbitrary dimensions and to spline filters, we can show that
\begin{equation}
(4 d \tau)^2 
\end{equation}
is such a bound. 
The key here is that the convolution with $b^{(\mathbf{0})}$ is bounded by $1$. 

We stop the iterations of the dual-based algorithm early because convergence of the outer proximal-gradient algorithm occurs even when the approximation of the application of the proximal operator is coarse \cite{villa_accelerated_2013}.

\subsection{Algorithm and Implementation}
Equipped with the gradient \eqref{eq:loglikelihood_derivative}, the proximal operator \eqref{eq:prox_hessian_schat} and the adaptive Lipschitz bounds \eqref{eq:lips_bound}, we setup an accelerated proximal-gradient descent algorithm with restarting scheme \cite{odonoghue_adaptive_2015}. The outline is shown in Algorithm \ref{algo}. We have also derived convolution-based expressions to accelerate the computation of the normalization of the integrals and the evaluation of the densities (Appendices B-C). 

\def\ctemp{c_{\text{temp}, k}}
\begin{algorithm}
 \caption{Optimization of $\func$ for $\opticoeffs$}
 \begin{algorithmic}[1]
 \renewcommand{\algorithmicrequire}{\textbf{Input:}}
 \renewcommand{\algorithmicensure}{\textbf{Output:}}
 \REQUIRE $\{\point_q\}$, $\mathcal{M}$, BCs, $\lambda$, $\epsilon_\text{tol}$
 \ENSURE $\opticoeffs$ 
 \STATE \textit{Initialization}: $c_{0} = c_{\text{temp}, 0} = -1$, $t_0=1$
  \FOR {$k = 0$ to $\text{max\_iter}-1$}
  \STATE $\tau = \lips(\ctemp)^{-1}$ (see \eqref{eq:lips_bound})
  \STATE $c_{k+1}=\prox_{\lambda\tau R}\left(\ctemp-\tau\mathbf{g}^* (\ctemp) \right)$ (see 
  
  \hfill \eqref{eq:loglikelihood_derivative},~ \eqref{eq:prox_hessian_schat})
  \IF {$\text{converged}(c_{k+1},c_{k}, \epsilon_\text{tol})$}
  \STATE \textbf{break}
  \ENDIF
  \STATE $t_{k+1}=\text{accelerate}(t_{k})$
  \STATE $\left( c_{k+1}, t_{k+1} \right) = \text{restart}(c_{k+1},c_{k},\ctemp,t_{k+1})$
  \STATE $c_{\text{temp}, k+1} = c_{k+1} + \left(c_{k+1} - c_{k} \right) \text{momentum}(t_{k+1}, t_{k})$
  \STATE  $\left( c_{k},  t_{k} \right) =  \left( c_{k+1},  t_{k+1} \right)$
  
  \ENDFOR
 \STATE $c^\star=c_{k+1}$
 \RETURN $c^\star$ 
 \end{algorithmic} 
  \label{algo}
 \end{algorithm}

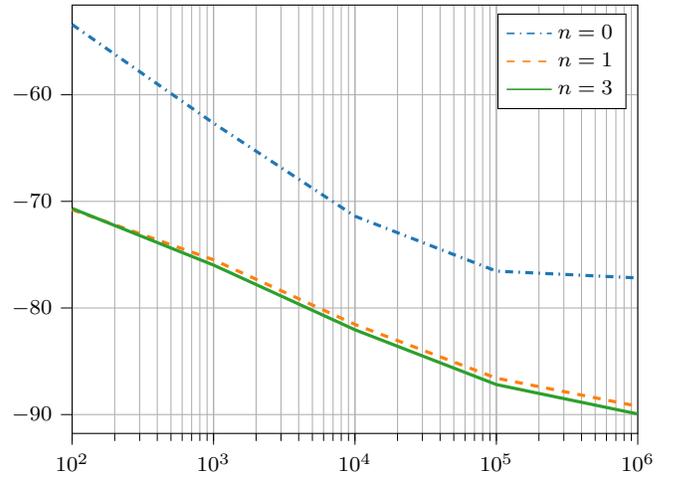
\begin{figure}
    \centering
\begin{tikzpicture}[every plot/.append style={very thick}]

\definecolor{darkgray176}{RGB}{176,176,176}
\definecolor{darkorange25512714}{RGB}{255,127,14}
\definecolor{forestgreen4416044}{RGB}{44,160,44}
\definecolor{steelblue31119180}{RGB}{31,119,180}
\def\ys{-4pt}
\begin{axis}[
title style={yshift=\ys,},
log basis x={10},
tick align=outside,
tick pos=left,
x grid style={darkgray176},
xmin=100, xmax=1000000,
xmode=log,
xtick style={color=black},
y grid style={darkgray176},
ymin=-91.7713842627766, ymax=-51.5830172716003,
ytick style={color=black},
title={\small MSE [dB] vs Number of Samples},
grid=both,
width=0.5\textwidth,
height=0.4\textwidth,
]
\addplot [semithick, steelblue31119180, dash dot]
table {%
100 -53.4097612257447
1000 -62.6667884002518
10000 -71.388064962509
100000 -76.5456033423892
1000000 -77.1840606446218
};
\addlegendentry{$n=0$}
\addplot [semithick, darkorange25512714, dashed]
table {%
100 -70.7792210990001
1000 -75.4839829830785
10000 -81.5433838737725
100000 -86.5856671096343
1000000 -89.2218224296417
};
\addlegendentry{$n=1$}
\addplot [semithick, forestgreen4416044]
table {%
100 -70.6621410929498
1000 -75.9774014326024
10000 -82.0462238290555
100000 -87.1859958072853
1000000 -89.9446403086322
};
\addlegendentry{$n=3$}
\end{axis}

\end{tikzpicture}
    \vspace{-10pt}
    \caption{MSE of the RDS estimate of the GG density as a function of the number of samples for spline degrees $n\in \{0,1,3\}$.}
    
    \label{fig:degree}
\end{figure}

In our implementation of Algorithm \ref{algo}, we chose the relative convergence criterion
\begin{equation}
\text{converged}(c_{k+1},c_{k}, \epsilon_{\text{tol}})= \mathbbm{1}_{<0}\left(\frac{\norm{c_{k+1}-c_{k}}}{\norm{c_{k}}}-\epsilon_{\text{tol}}\right)
\end{equation}
with tolerance $\epsilon_{\text{tol}}$. The acceleration was chosen as the standard
\begin{equation}
\text{accelerate}(t_k)=\frac{1+\sqrt{1+4t_k^2}}{2},
\end{equation}
together with
\begin{equation}
\text{momentum}(t_{k+1},t_{k})=\frac{t_k-1}{t_{k+1}}.
\end{equation}
The restarting scheme was implemented as
\begin{align}
&\text{restart}(c_{k+1},c_{k},\ctemp,t_{k+1})= \nonumber \\
& \hspace*{-0.1cm} \begin{cases}
(c_{k+1}, t_{k+1}) \text{ if } \langle\ctemp-c_{k+1},c_{k+1}-c_{k}\rangle_{\mathcal{C}}\,<0 \\
 (c_{k}, 1) \text{ otherwise}.
\end{cases}
\end{align}
The algorithm is initialized with negative coefficients.

The library resulting from this work will be available online\footlabel{note1}{\href{https://github.com/AleixBP/rdsplines}{github.com/AleixBP/rdsplines} }. It is written in general dimension and based exclusively on NumPy and SciPy; it also offers GPU support through their CuPy counterparts.

\def\our{RDS }

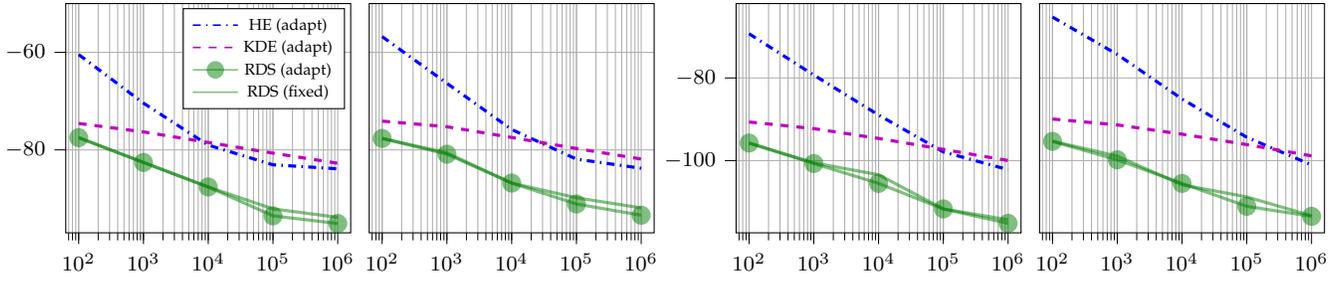
\begin{figure*}
    \centering
\begin{tikzpicture}[every plot/.append style={very thick}]
\def\imwidth{.3\textwidth}\def\hs{25pt}\def\vs{10pt}

\definecolor{darkcyan30160135}{RGB}{30,160,135}
\definecolor{darkcyan35137141}{RGB}{35,137,141}
\definecolor{darkgray176}{RGB}{176,176,176}
\definecolor{darkslateblue6077138}{RGB}{60,77,138}
\definecolor{darkslateblue6664133}{RGB}{66,64,133}
\definecolor{darkslateblue7142121}{RGB}{71,42,121}
\definecolor{gold21822624}{RGB}{218,226,24}
\definecolor{gold25323136}{RGB}{253,231,36}
\definecolor{indigo68184}{RGB}{68,1,84}
\definecolor{mediumseagreen33166133}{RGB}{33,166,133}
\definecolor{mediumseagreen71192110}{RGB}{71,192,110}
\definecolor{mediumseagreen9620196}{RGB}{96,201,96}
\definecolor{steelblue46108142}{RGB}{46,108,142}
\definecolor{teal44114142}{RGB}{44,114,142}
\definecolor{darkgray176}{RGB}{176,176,176}
\definecolor{steelblue31119180}{RGB}{31,119,180}

\definecolor{darkgray176}{RGB}{176,176,176}
\definecolor{darkviolet1910191}{RGB}{191,0,191}
\definecolor{green01270}{RGB}{0,127,0}
\def\ys{-4pt}

\begin{groupplot}[group style={group size=4 by 1,horizontal sep=\hs}, width=.295\textwidth] 
\nextgroupplot[
title style={yshift=\ys,},
log basis x={10},
tick align=outside,
tick pos=left,
x grid style={darkgray176},
xmin=63.0957344480193, xmax=1584893.19246111,
xmode=log,
xtick style={color=black},
y grid style={darkgray176},
ymin=-97, ymax=-50,
ytick style={color=black},
title={\small MSE [dB] vs Number of Samples},
legend style={nodes={scale=0.75, transform shape}},
grid=both
]
\addplot [semithick, blue, dash dot]
table {%
100 -60.4543105305888
1000 -70.4139756425328
10000 -79.0169829047844
100000 -83.0509769614668
1000000 -83.871905521225
};
\addlegendentry{HE (adapt)}
\addplot [semithick, darkviolet1910191, dashed]
table {%
100 -74.5744544527352
1000 -76.3080882568629
10000 -78.476631949646
100000 -80.663354572997
1000000 -82.7424241754299
};
\addlegendentry{KDE (adapt)}
\addplot [semithick, green01270, mark=*, mark size=3, mark options={solid},opacity=0.5]
table {%
100 -77.4839012735886
1000 -82.572676554016
10000 -87.5966737174136
100000 -93.5205375295942
1000000 -95.0879737005883
};
\addlegendentry{RDS (adapt)}
\addplot [semithick, green01270,opacity=0.5]
table {%
100 -77.4839012735886
1000 -82.572676554016
10000 -87.5966737174136
100000 -92.0912413791095
1000000 -93.8211999517813
};
\addlegendentry{RDS (fixed)}

\nextgroupplot[
log basis x={10},
tick align=outside,
tick pos=left,
x grid style={darkgray176},
xmin=63.0957344480193, xmax=1584893.19246111,
xmode=log,
xtick style={color=black},
y grid style={darkgray176},
ymin=-97, ymax=-50,
ytick style={color=black},
grid=both,
ymajorticks=false,
xshift=-18pt,
]
\addplot [semithick, blue, dash dot]
table {%
100 -56.7726209796978
1000 -66.3447217518729
10000 -75.8197543382415
100000 -81.8873059553308
1000000 -83.7631679880448
};
\addplot [semithick, green01270,opacity=0.5]
table {%
100 -77.6454448389803
1000 -80.5973262438929
10000 -86.7985708430295
100000 -89.8115215057121
1000000 -91.890313481324
};
\addplot [semithick, green01270, mark=*, mark size=3, mark options={solid},opacity=0.5]
table {%
100 -77.6455069306996
1000 -80.8781668342673
10000 -86.7985708430295
100000 -91.086994607355
1000000 -93.3945461533776
};
\addplot [semithick, darkviolet1910191, dashed]
table {%
100 -74.1008456321932
1000 -75.2471904021276
10000 -77.4459611939551
100000 -79.7137464412655
1000000 -81.8711956584048
};
\nextgroupplot[
title style={yshift=\ys,},
log basis x={10},
tick align=outside,
tick pos=left,
x grid style={darkgray176},
xmin=63.0957344480193, xmax=1584893.19246111,
xmode=log,
xtick style={color=black},
y grid style={darkgray176},
ymin=-117.475807151407, ymax=-62.0075292921093,
ytick style={color=black},
grid=both,
xshift=6pt,
]
\addplot [semithick, blue, dash dot]
table {%
100 -69.3015419220774
1000 -79.3014061465167
10000 -88.9626360708045
100000 -97.8994544304174
1000000 -102.187363423225
};
\addplot [semithick, green01270,opacity=0.5]
table {%
100 -95.7447400604081
1000 -100.544295412967
10000 -103.426783590897
100000 -111.698522448902
1000000 -114.317889990202
};
\addplot [semithick, green01270, mark=*, mark size=3, mark options={solid},opacity=0.5]
table {%
100 -95.7537042938319
1000 -100.699198344083
10000 -105.519606996984
100000 -111.698522448902
1000000 -115.181794521439
};
\addplot [semithick, darkviolet1910191, dashed]
table {%
100 -90.6570015255917
1000 -92.2643476093019
10000 -94.637622585782
100000 -97.2822946227357
1000000 -99.9963242971954
};
\nextgroupplot[
log basis x={10},
tick align=outside,
tick pos=left,
x grid style={darkgray176},
xmin=63.0957344480193, xmax=1584893.19246111,
xmode=log,
xtick style={color=black},
y grid style={darkgray176},
ymin=-117.475807151407, ymax=-62.0075292921093,
ytick style={color=black},
grid=both,
xshift=8pt,
ymajorticks=false,
xshift=-26pt,
]
\addplot [semithick, blue, dash dot]
table {%
100 -65.2651500899804
1000 -74.3127805379968
10000 -85.0870998143597
100000 -94.3544504386836
1000000 -101.043239526166
};
\addplot [semithick, green01270,opacity=0.5]
table {%
100 -95.3076347422768
1000 -98.9111263759325
10000 -105.853396225056
100000 -108.781642763499
1000000 -113.35392056276
};
\addplot [semithick, green01270, mark=*, mark size=3, mark options={solid},opacity=0.5]
table {%
100 -95.350502704987
1000 -99.7981501561974
10000 -105.519606996984
100000 -111.064347639549
1000000 -113.480511599309
};
\addplot [semithick, darkviolet1910191, dashed]
table {%
100 -89.9370189524707
1000 -91.3691319454832
10000 -93.6029458173092
100000 -96.1313396322715
1000000 -98.8606264530389
};
\end{groupplot}
\end{tikzpicture}
    \vspace{-10pt}
    \caption{MSE of the RDS estimate of the UGL density as a function of the number of samples for (left to right) sU, sS in 2D, and  sU, sS in 3D, respectively. RDS (adapt) runs a small grid-search (three values) for the best $\lambda$ at each number of samples, whereas (fixed) uses the same, fixed $\lambda$ throughout all the range. }
   \label{fig:mse_graphs}
\end{figure*}

\section{Experiments} \label{sec:experiments}
We tested our framework on standard theoretical distributions and on PET sinograms. We evaluated the accuracy of each method by comparing the resulting estimator $\hat{\density}$ with the ground-truth density $\density$ according to
\begin{equation}
\text{MSE} (\density,\hat{\density}) =  10 \log_{10} \left( \frac{1}{|\mathcal{M}_s|} \norm{\samp_{\mathcal{M}_s}\{  \density \} - \samp_{\mathcal{M}_s}\{  \hat{\density}\}}_F^2 \right),
\end{equation}
which involves the mean-squared error (MSE) on some fine grid $\mathcal{M}_s$.
Hereafter, we refer to our framework as \our for regularized-density splines.

To simulate samples $\point \sim \nu$ from the observed density, we first take samples $\mathbf{p} \sim \dens$ from the underlying density of interest. We then thin them according to the probability $\sensi/\max(\sensi)$. More precisely, we keep the sample $\mathbf{p}$ if $\sensi(\mathbf{p})/\max(\sensi) > u$, where $u \sim U(0,1)$ comes from a uniform distribution on the unit interval. This is known as rejection sampling.

\subsection{Standard Distributions}
 We considered two different distributions $\nu$
 in the role of 
 standard distributions: UGL and GG. UGL is a compound distribution made of the sum of a uniform distribution, a Gaussian distribution, and a Laplacian distribution (Figure \ref{fig:samples}, leftmost). We extended it to $n$D by considering independent Laplacians along each dimension. GG is the sum of two Gaussians. (Find the exact list of all parameters online\footref{note1}.) The domain of the distributions was taken to be periodic.
 We tested them for dimensions $d=2$ and $d=3$. The sensitivity maps were chosen among $\xi(\point)=1$ and $\xi(\point) = \sin^2(x_1/T+\phi) + \epsilon$ with a phase $\phi$, a period $T$, and $\epsilon=10^{-3}$ (Figure \ref{fig:samples}, middle). We call these sensitivity maps sU and sS, respectively. 
 
 We chose the grid as $\mathcal{M} = \prod_{k=1}^d\listintzero{N_k-1}$ and set the basis according to the periodic domain as
\begin{equation}\label{eq:periodic_bcs}
            \varphi(\point) = \prod_{k=1}^d 
            \beta^n(\point_k \bmod \left(N_k-1 \right)).
\end{equation}
\tc{Note the ``lack'' of a basis at $N_k$ because it is the basis at $0$.} 
The number of samples that we considered ranged from $10^2$ to $10^6$.

\subsection{Preliminary Analysis}

\subsubsection{Stepsize Adapatation}
We have validated the importance of adapting the Lipschitz constant at each iteration of the optimization problem. To this end, we have compared the convergence of the algorithm for two schemes: one where the Lipschitz ``constant" is updated, which we refer to as $\lips(\ctemp)$; and another where it remains constant throughout the iterations after reinitialization at $\coeffs_1$, which we refer to as $\lips(\coeffs_1)$. In both cases, the bound is computed as per \eqref{eq:lips_ineq}.
In Figure \ref{fig:lipschitz}, one can see how the evolution of $\lips(\ctemp)$ leads to fast convergence.
Conversely, when using a non-varying $\lips(\coeffs_1)$ we have observed that sometimes the optimization diverges straight away for certain initializations, and some other times strong oscillations arise after some descent.

\subsubsection{Degree} In our experiments, we observed that the biggest improvement comes when moving from splines of degree $0$ to splines of degree $1$. Splines of degree $3$ did not increase the accuracy that much (Figure \ref{fig:degree}) in comparison to their much higher computational cost. This is in line with our previous experience of working with splines. %
Accordingly, we set $n=1$ henceforth.

\subsubsection{Effect of the Regularization}
We first checked what kind of sparsity the linear splines induce along the axes. We used a Laplacian distribution along the first axis for $d=2$ for this test. We observed that the minimization of the nuclear norm resulted in a sparse Hessian, which here acted as a surrogate for knot sparsity along the first axis (Figure \ref{fig:sparsity}).

In Figure \ref{fig:lambda_vs_mse}, we looked at the sensitivity of the RDS estimates to the regularization parameter.
For a given number of samples, the optimal $\lambda$ did not depend on the grid size over the range tested (Figure \ref{fig:lambda_vs_mse}, vertical lines). The number of samples affected the optimal $\lambda$ in a predictable way: multiplying the number of samples by factors of $10$ required the multiplication of the optimal $\lambda$ by factors of $2$ (Figure \ref{fig:lambda_vs_mse}, vertical lines). Above a certain grid size (relative to the number of samples), the dependency of the MSE on $\lambda$ reached a plateau at the minimum error (Figure \ref{fig:lambda_vs_mse}). This phase-transition behavior is reminiscent of sparse optimization and allows leeway in choosing an optimal $\lambda$. We also observed the diminishing returns in MSE as the scale of the grid or the number of samples increase.

\subsection{Error and Speed in Standard Distributions} \label{sec:example_density}
We compared RDS to KDE and HE. The bandwidth of HE was adjusted
according to the maximum among the Freedman-Diaconis’ and Sturges’
estimators, whereas that of KDE follows Scott’s rule. The implementation of KDE and HE were taken from SciPy. We chose a Gaussian distribution (radial basis function) for the kernel of the KDE. 

\subsubsection{Quantitative Assessment} We evaluated the MSE as a function of the number of samples. RDS outperformed HE and KDE in all cases and over the whole range of samples (Figure \ref{fig:mse_graphs}, UGL, $d \in \{2,3\}$, sU, and sS). For the number of samples relevant to imaging modalities ($\geq 10^5$), the improvement amounted to an order of magnitude (Figure \ref{fig:mse_graphs}, leftmost).

RDS was robust to the sS sensitivity, only losing around $\unit{0.1}{dB}$ throughout the sample range with respect to sU (Figure \ref{fig:mse_graphs}, middle). KDE was also quite robust ($\unit{0.6}{dB}$ lost). The irregular weights in KDE were compensated by over-smoothing. This was to the detriment of the qualitative value of the estimate because modes blended together. In contrast, HE became much worse---its MSE decreased by around $\unit{4}{dB}$---but remained sharp.

In 3D, the relative improvement with RDS was even larger, maxing out at \unit{15}{dB} for $10^6$ samples (Figure \ref{fig:mse_graphs}, rightmost). From 2D to 3D, HE became worse in comparison to KDE because hyper-bins quickly become empty due to the curse of dimensionality.

To further explore the robustness of the method to $\lambda$, we compared two versions of RDS. One where $\lambda$ was adjusted at each number of samples according to the best MSE among five values (corresponding to five orders of magnitude). And another one where $\lambda$ was fixed to the optimal value found for $10^2$ samples. They both performed similarly (Figure \ref{fig:mse_graphs}).

\begin{figure*}
    \centering
    \input{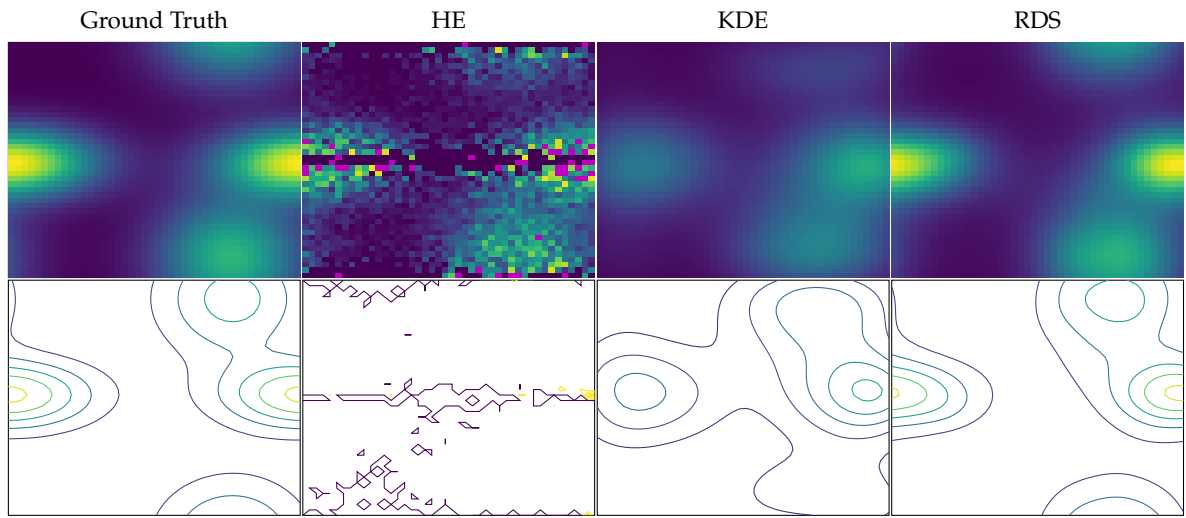}
    \vspace{-6pt}
    \caption{Estimates of the ground truth by HE, KDE, and RDS (left to right) from $5 \cdot 10^3$ samples under GG-sS. MSEs [dB]: $-61$, $-71$, $-84$, respectively. Bottom row are contour plots of top row.}
    
    \label{fig:panels_gaussians}
\end{figure*}

\subsubsection{Qualitative Assessment} 
\begin{figure*}
    \centering
    \input{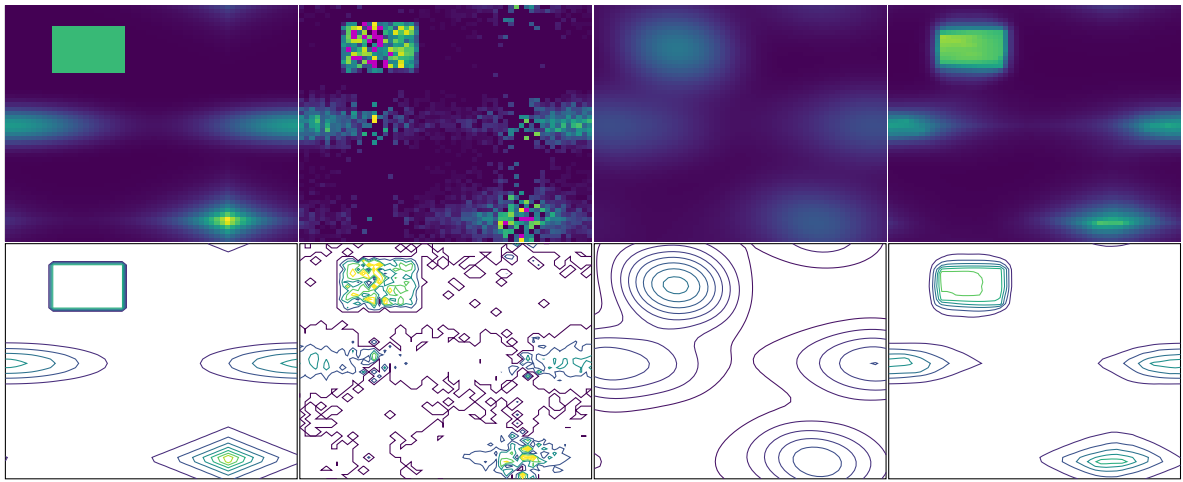}
    \vspace{-6pt}
    \caption{Estimates of the ground truth by HE, KDE, and RDS (left to right) from $5 \cdot 10^3$ samples under UGL-sS. MSEs [dB]: $-73$, $-77$, $-85$, respectively. Bottom row are contour plots of top row.}
    \label{fig:mse_panels_yesweight4}
    
\end{figure*}

\begin{figure*}
    \centering
    \input{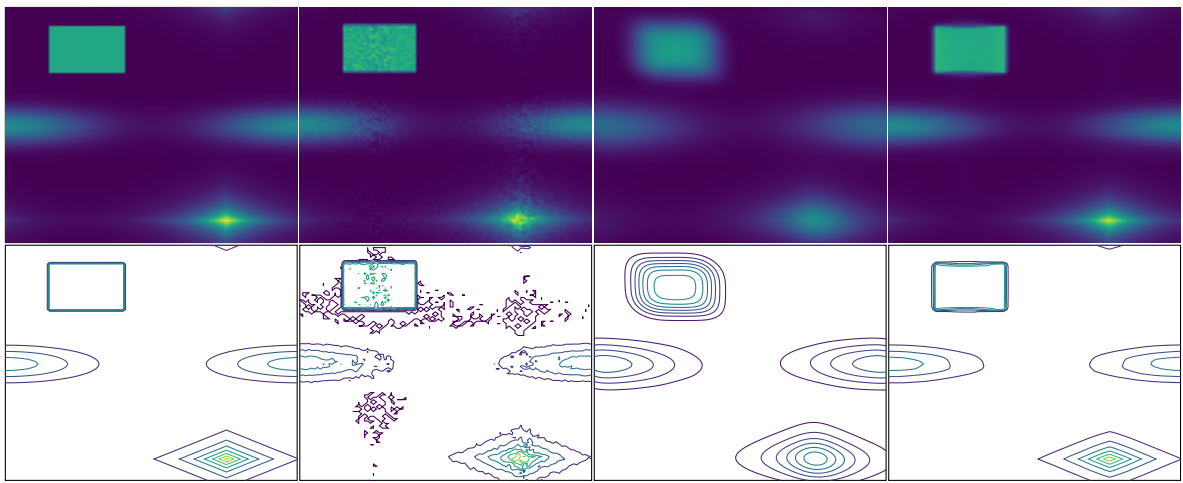}
    \vspace{-6pt}
    \caption{Estimates of the ground truth by HE, KDE, and RDS (left to right) from $5 \cdot 10^5$ samples under UGL-sS. MSEs [dB]: $-83$, $-81$, $-91$, respectively. Bottom row are contour plots of top row.}
    \label{fig:mse_panels_yesweight6}
    
\end{figure*}

To better favor the Gaussian kernel of the KDE, we first assessed GG at $5 \cdot 10^3$ samples (Figure \ref{fig:panels_gaussians}). We observe that RDS adapted well to regions of different underlying variance even under heterogeneous sensitivity. It also compensates well for less sensitive zones, where data are scarce. Conversely, KDE compensates for the lack of samples by over-smoothing most of the domain, while HE is unable to correct for the sensitivity. Similar observations apply to the estimation of the UGL distribution (Figures \ref{fig:mse_panels_yesweight4} and \ref{fig:mse_panels_yesweight6}, $5 \cdot 10^3$ and $5 \cdot 10^5$ samples, respectively).

\subsubsection{Speed} 
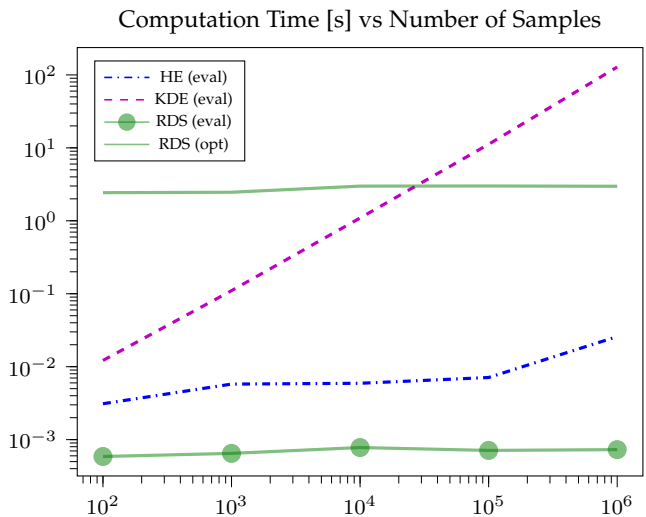
\begin{figure}
    \centering
\begin{tikzpicture}[every plot/.append style={very thick}]

\definecolor{darkgray176}{RGB}{176,176,176}
\definecolor{darkviolet1910191}{RGB}{191,0,191}
\definecolor{green01270}{RGB}{0,127,0}
\def\ys{-4pt}
\begin{axis}[
title style={yshift=\ys,},
log basis x={10},
log basis y={10},
tick align=outside,
tick pos=left,
x grid style={darkgray176},
xmin=63.0957344480193, xmax=1584893.19246111,
xmode=log,
xtick style={color=black},
y grid style={darkgray176},
ymin=0.000317718079058116, ymax=236.972027887546,
ymode=log,
ytick style={color=black},
title={\small Computation Time [s] vs Number of Samples},
legend pos=north west,
legend style={nodes={scale=0.75, transform shape}},
width=0.5\textwidth,
height=0.4\textwidth,
]
\addplot [semithick, blue, dash dot]
table {%
100 0.0031063270568847654
1000 0.005773429870605469
10000 0.005923948287963867
100000 0.00710965633392334
1000000 0.02583645820617676
};
\addlegendentry{HE (eval)}
\addplot [semithick, darkviolet1910191, dashed]
table {%
100 0.0121712684631348
1000 0.110234498977661
10000 1.09203290939331
100000 11.2049670219421
1000000 128.161686658859
};
\addlegendentry{KDE (eval)}

\addplot [semithick, green01270, mark=*, mark size=3, mark options={solid},opacity=0.5]
table {%
100 0.00058746337890625
1000 0.000648021697998047
10000 0.000779390335083008
100000 0.000711679458618164
1000000 0.000729799270629883
};
\addlegendentry{RDS (eval)}

\addplot [semithick, green01270,opacity=0.5]
table {%
100 2.42947006225586
1000 2.46265625953674
10000 2.98916125297546
100000 3.00050091743469
1000000 2.97487425804138
};

\addlegendentry{RDS (opt)}

\end{axis}

\end{tikzpicture}
    \vspace{-10pt}
    \caption{Computation time needed to evaluate/optimize the three methods in 2D. PET applications typically involve $10^6$ samples.}
    
    \label{fig:mse_comparison_noweight_time}
\end{figure}

Another advantage of RDS over KDE is a reduced computation time. RDS took about $\unit{1}{\second}$ to optimize and $\unit{10^{-3}}{\second}$ to evaluate in 2D. In our tests, this did not depend on the number of samples (Figure \ref{fig:mse_comparison_noweight_time}). The cost of optimization for KDE is negligible with respect to the cost of its evaluation. A single evaluation went from $\unit{10^{-2}}{\second}$ to $\unit{10^{2}}{\second}$ in the range of samples we considered. This meets the optimization time of RDS for as few as $10^4$ samples. The computation time for HE also increased with the number of samples, but stayed low overall. 
Consequently, RDS is particulary favorable in the presence of many samples or when repeated evaluations are needed. For instance, we expect substantial computational savings for imaging modalities such as PET since they routinely deal with over $10^6$ samples.

The results in 3D were consistent with those in 2D. RDS was still independent of the number of samples. The optimization of RDS in 3D was an order of magnitude slower than in 2D, whereas evaluations were only 1.2 times slower. KDE evaluations were an order of magnitude slower than their 2D counterparts.

\section{Applications to Imaging}

\label{sec:applications}

    Super-resolution microscopy and, more recently, single-photon emission computed tomography and PET work with low photon counts. The randomness of both electron excitation and nuclear decay results in photon emissions that are well described by inhomogeneous Poisson processes with spatially varying intensities $\check{\pi}$.
    Once $N_\text{photons}$ have eventually been detected, however, we argue that the problem boils down to that of the estimation of the pdf $\density = \check{\pi}/\int_\sinodomain \check{\pi}$ 
     of the source that generates the point clouds, where $\int_\sinodomain \check{\pi}=N_\text{photons}$. We will illustrate these ideas with PET examples.
     
     \subsection{Background on PET}
    PET reconstructs images from photon pairs that are emitted in proportion to the spatial distribution of radioactivity in the sample. The lines joining these pairs are called lines of response (LOR) and constitute the measurements \cite{defrise_exact_1997}. The sensitivity of the scanner to the LORs is corrected by pointwise division with respect to a reference scan. 
As it turns out, in some state-of-the-art scanners, detectors are so small that most LORs accrue a single pair of photons ($\sim 0.95\%$) or none at all ($\sim 99\%$) instead of hundreds (Figure~\ref{fig:pet}a)~\cite{iacobucci_monolithic_2021, iacobucci_efficiency_2022,yamamoto_development_2016, cadoux_100mupet_2023}. As a consequence, we argue that the interpolation in typical PET rebinning could be better regarded from the perspective of DE. The correction of the heterogeneous sensitivity of these scanners is another challenge as it can span a whole order of magnitude (Figure~\ref{fig:pet}b).

The set of measurements is called a tomographic sinogram. There, we regard the pairs of angle and ``distance" coordinates of LORs as samples $\point =(\theta, s)$ set on $\sinodomain = [0, \pi) \times [-\rho, \rho]$, where $\rho$ is the radius of the field of view of the scanner---no LOR can exist outside of it. This also means that $\density(\theta, \rho)=\density(\theta, -\rho)=0$ for all $\theta \in [0, \pi)$. The angular BC is periodic and the distance BC is constant. Each sample corresponds to a LOR connecting a pair of antiparallel photons detected by the scanner at (roughly) the same time but different detectors.

\subsection{PET Experiments}
We tested RDS against KDE and HE in the context of sinogram rebinning for an experimental small-animal PET scanner. The detectors of this scanner are very small~\cite{iacobucci_monolithic_2021, iacobucci_efficiency_2022,yamamoto_development_2016}. We also evaluated the quality of the reconstructions from the sinograms. The bandwidths for KDE and HE were chosen as in section \ref{sec:example_density}. The typical approach to PET rebinning using function interpolation (as opposed to \de) was omitted because it did not apply to such a sparse data regime. \de was more appropriate because the number of samples was small in comparison to the size of the detectors (Figure \ref{fig:pet}a).

We first assessed the sensitivity $\sensi$ of the scanner. Measurements in the sinogram domain $\sinodomain$ indicated that the chance of detecting a LOR was very heterogeneous (Figure \ref{fig:pet}b). We registered a difference of up to an order of magnitude between the highest and lowest sensitivity. This had a sizeable impact on the resulting sinograms (Figure \ref{fig:pet}c-d).

We then assumed the underlying concentration of radioactive material (radiotracer) to be the sum of four Gaussians (data not shown, see \cite[Figures~3-4]{boquet-pujadas_pet_2023}). The reasons were twofold: for a comparison that favors KDE, and for ease of the interpretation of the sinogram. PET being a tomographic modality, Gaussianity was preserved over $s$ on the resulting sinogram. The PET acquisition was simulated for scanning times corresponding to $10^2$ to $10^6$ samples. The resulting MSEs behaved almost identically to those in Figure~\ref{fig:mse_graphs}. At the PET-relevant mark of $10^6$ samples, RDS performed more than an order of magnitude better than KDE and HE. Our qualitative observations were also the same as for the previous analysis on standard distributions.

\begin{figure*}
    \centering
    \begin{tikzpicture}
\def\imwidth{.335\textwidth}
\def\hs{4pt}
\definecolor{darkgray176}{RGB}{176,176,176}
\def\twidth{.3\textwidth}
\def\hwidth{.268\textwidth}
\begin{groupplot}[group style={group size=4 by 1,horizontal sep=\hs}, ticks=none,
width=\imwidth,
title style={yshift=-4pt}]
\nextgroupplot[
hide x axis,
hide y axis,
point meta max=0.000217391304347826,
point meta min=0,
tick align=outside,
tick pos=left,
x grid style={darkgray176},
 xmin=-14., xmax=14., 
 ymin=-12.5, ymax=12.5, 
xtick style={color=black},
y grid style={darkgray176},
ytick style={color=black},
height=\hwidth,
width=\twidth,
]
\addplot graphics [includegraphics cmd=\pgfimage,xmin=-17.5, xmax=17.5, ymin=-17.5, ymax=17.5] {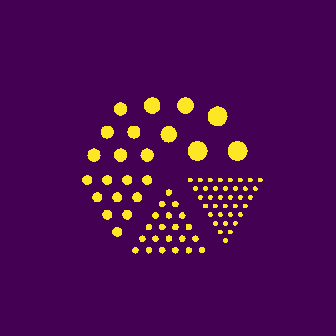};

\nextgroupplot[
hide x axis,
hide y axis,
tick align=outside,
tick pos=left,
x grid style={darkgray176},
 xmin=-14., xmax=14., 
 ymin=-12.5, ymax=12.5, 
xtick style={color=black},
y grid style={darkgray176},
ytick style={color=black},
height=\hwidth,
width=\twidth,
]
\addplot graphics [includegraphics cmd=\pgfimage,xmin=-17.5, xmax=17.5, ymin=-17.5, ymax=17.5] {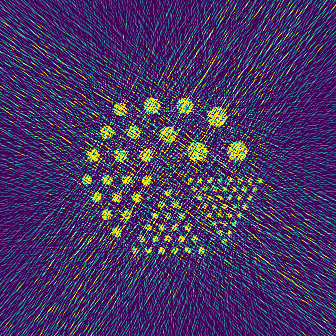};

\nextgroupplot[
hide x axis,
hide y axis,
tick align=outside,
tick pos=left,
x grid style={darkgray176},
 xmin=-14., xmax=14., 
 ymin=-12.5, ymax=12.5, 
xtick style={color=black},
y grid style={darkgray176},
ytick style={color=black},
height=\hwidth,
width=\twidth,
]
\addplot graphics [includegraphics cmd=\pgfimage,xmin=-17.5, xmax=17.5, ymin=-17.5, ymax=17.5] {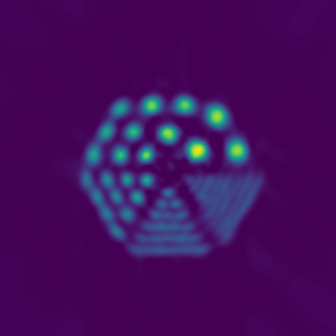};

\nextgroupplot[
hide x axis,
hide y axis,
tick align=outside,
tick pos=left,
x grid style={darkgray176},
xtick style={color=black},
y grid style={darkgray176},
 ymin=-12.5, ymax=12.5, 
 xmin=-14., xmax=14., 
ytick style={color=black},
height=\hwidth,
width=\twidth,
]
\addplot graphics [includegraphics cmd=\pgfimage,xmin=-17.5, xmax=17.5, ymin=-17.5, ymax=17.5] {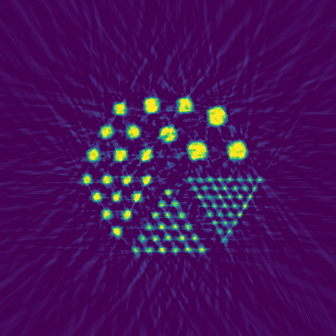};

\end{groupplot}

\end{tikzpicture}
    \vspace{-4pt}
    \caption{FBP reconstruction of a Derenzo phantom (leftmost) after resampling the sinogram (Figure \ref{fig:pet}) with (left to right) HE, KDE, and RDS. MSEs [dB]: -184, -204, -211, respectively.}
    \label{fig:derenzo}
\end{figure*}

The improved sinogram estimation of RDS translated into better MSE in the image domain upon reconstruction with the filtered back-projection (FBP). This was the case in a Derenzo phantom, the most common benchmark for PET systems (Figure \ref{fig:derenzo} and \ref{fig:pet}d). RDS managed to reduce the streak artifacts caused by the FBP without suffering the over-smoothing of KDE. Since the FBP requires that the evaluations be performed on a grid, the convolution properties of RDS also offer an advantage in performance.

\subsubsection{Data and Phantoms}
\tc{We next tested the algorithm under more practical~conditions.
}

\begin{figure*}
    \centering
\begin{tikzpicture}

\definecolor{darkgray176}{RGB}{176,176,176}

\def\ys{-4pt}
\def\imwidth{.3\textwidth}\def\hs{1pt}\def\vs{1pt}
\begin{groupplot}[group style={group size=4 by 5,horizontal sep=\hs, vertical sep=\vs}, ticks=none, width=\imwidth]

\nextgroupplot[
tick align=outside,
tick pos=left,
x grid style={darkgray176},
xmin=-0.5, xmax=811.5,
xtick style={color=black},
y dir=reverse,
y grid style={darkgray176},
ymin=-0.5, ymax=491.5,
ytick style={color=black},
title={\small Ground Truth},
title style={yshift=\ys,},
ylabel={\small Brain Phantom},
ylabel style={yshift=-\ys,},
]
\addplot graphics [includegraphics cmd=\pgfimage,xmin=-0.5, xmax=811.5, ymin=491.5, ymax=-0.5] {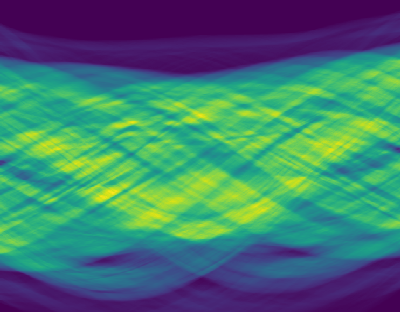};

\nextgroupplot[
hide x axis,
hide y axis,
tick align=outside,
tick pos=left,
x grid style={darkgray176},
xmin=-0.5, xmax=811.5,
xtick style={color=black},
y dir=reverse,
y grid style={darkgray176},
ymin=-0.5, ymax=491.5,
ytick style={color=black},
title={\small HE},
title style={yshift=\ys,},
]
\addplot graphics [includegraphics cmd=\pgfimage,xmin=-0.5, xmax=811.5, ymin=491.5, ymax=-0.5] {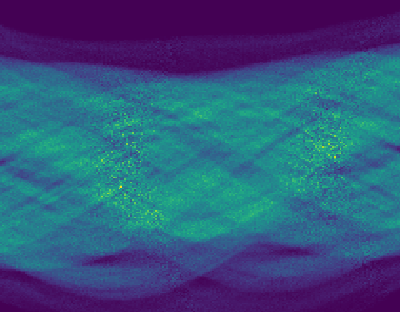};

\nextgroupplot[
hide x axis,
hide y axis,
tick align=outside,
tick pos=left,
x grid style={darkgray176},
xmin=-0.5, xmax=811.5,
xtick style={color=black},
y dir=reverse,
y grid style={darkgray176},
ymin=-0.5, ymax=491.5,
ytick style={color=black},
title={\small KDE},
title style={yshift=\ys,},
]
\addplot graphics [includegraphics cmd=\pgfimage,xmin=-0.5, xmax=811.5, ymin=491.5, ymax=-0.5] {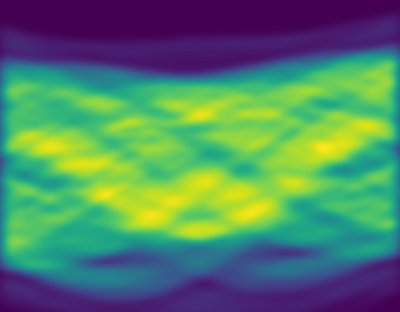};

\nextgroupplot[
hide x axis,
hide y axis,
point meta max=1.15928660492868e-05,
point meta min=1.15571540001439e-08,
tick align=outside,
tick pos=left,
x grid style={darkgray176},
xmin=-0.5, xmax=811.5,
xtick style={color=black},
y dir=reverse,
y grid style={darkgray176},
ymin=-0.5, ymax=491.5,
ytick style={color=black},
title={\small RDS},
title style={yshift=\ys,},
]
\addplot graphics [includegraphics cmd=\pgfimage,xmin=-0.5, xmax=811.5, ymin=491.5, ymax=-0.5] {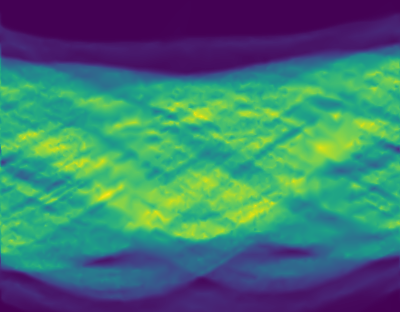};

\nextgroupplot[
tick align=outside,
tick pos=left,
x grid style={darkgray176},
xmin=-0.5, xmax=811.5,
xtick style={color=black},
y dir=reverse,
y grid style={darkgray176},
ymin=-0.5, ymax=491.5,
ytick style={color=black},
ylabel={\small Brain Phantom},
ylabel style={yshift=-\ys,},
]
\addplot graphics [includegraphics cmd=\pgfimage,xmin=-0.5, xmax=811.5, ymin=491.5, ymax=-0.5] {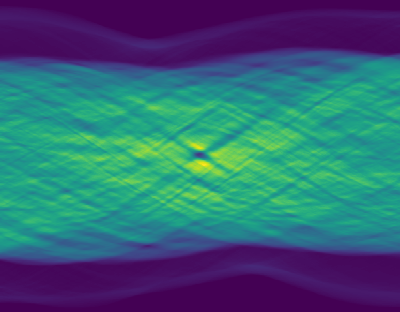};

\nextgroupplot[
hide x axis,
hide y axis,
tick align=outside,
tick pos=left,
x grid style={darkgray176},
xmin=-0.5, xmax=811.5,
xtick style={color=black},
y dir=reverse,
y grid style={darkgray176},
ymin=-0.5, ymax=491.5,
ytick style={color=black},
]
\addplot graphics [includegraphics cmd=\pgfimage,xmin=-0.5, xmax=811.5, ymin=491.5, ymax=-0.5] {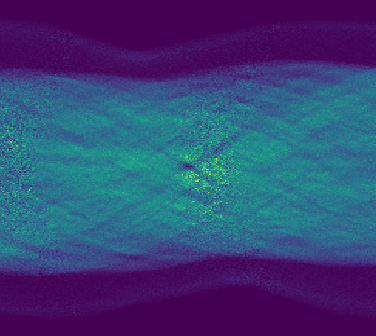};

\nextgroupplot[
hide x axis,
hide y axis,
tick align=outside,
tick pos=left,
x grid style={darkgray176},
xmin=-0.5, xmax=811.5,
xtick style={color=black},
y dir=reverse,
y grid style={darkgray176},
ymin=-0.5, ymax=491.5,
ytick style={color=black},
]
\addplot graphics [includegraphics cmd=\pgfimage,xmin=-0.5, xmax=811.5, ymin=491.5, ymax=-0.5] {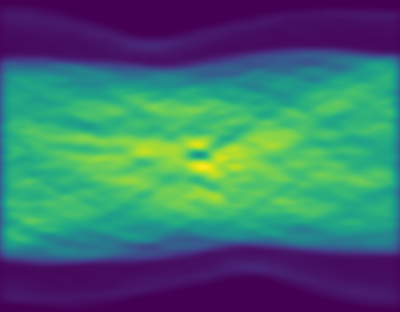};

\nextgroupplot[
hide x axis,
hide y axis,
point meta max=1.15928660492868e-05,
point meta min=1.15571540001439e-08,
tick align=outside,
tick pos=left,
x grid style={darkgray176},
xmin=-0.5, xmax=811.5,
xtick style={color=black},
y dir=reverse,
y grid style={darkgray176},
ymin=-0.5, ymax=491.5,
ytick style={color=black},
]
\addplot graphics [includegraphics cmd=\pgfimage,xmin=-0.5, xmax=811.5, ymin=491.5, ymax=-0.5] {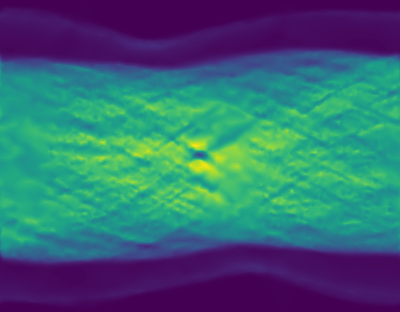};

\nextgroupplot[
tick align=outside,
tick pos=left,
x grid style={darkgray176},
xmin=-0.5, xmax=811.5,
xtick style={color=black},
y dir=reverse,
y grid style={darkgray176},
ymin=-0.5, ymax=491.5,
ytick style={color=black},
ylabel={\small Brain Phantom},
ylabel style={yshift=-\ys,},
]
\addplot graphics [includegraphics cmd=\pgfimage,xmin=-0.5, xmax=811.5, ymin=491.5, ymax=-0.5] {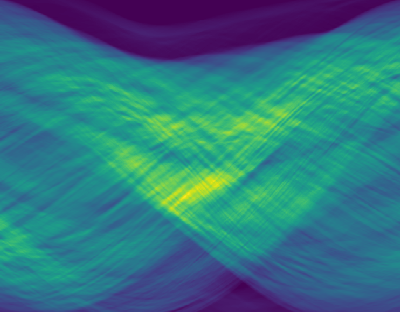};

\nextgroupplot[
hide x axis,
hide y axis,
tick align=outside,
tick pos=left,
x grid style={darkgray176},
xmin=-0.5, xmax=811.5,
xtick style={color=black},
y dir=reverse,
y grid style={darkgray176},
ymin=-0.5, ymax=491.5,
ytick style={color=black},
]
\addplot graphics [includegraphics cmd=\pgfimage,xmin=-0.5, xmax=811.5, ymin=491.5, ymax=-0.5] {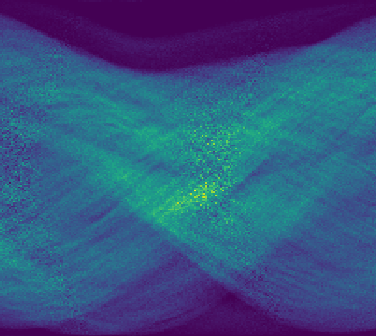};

\nextgroupplot[
hide x axis,
hide y axis,
tick align=outside,
tick pos=left,
x grid style={darkgray176},
xmin=-0.5, xmax=811.5,
xtick style={color=black},
y dir=reverse,
y grid style={darkgray176},
ymin=-0.5, ymax=491.5,
ytick style={color=black},
]
\addplot graphics [includegraphics cmd=\pgfimage,xmin=-0.5, xmax=811.5, ymin=491.5, ymax=-0.5] {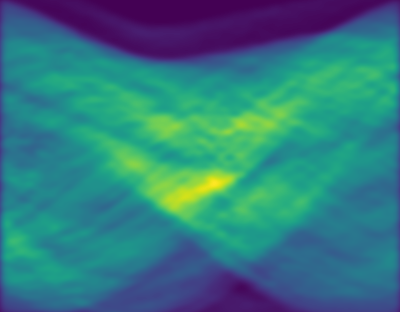};

\nextgroupplot[
hide x axis,
hide y axis,
point meta max=1.15928660492868e-05,
point meta min=1.15571540001439e-08,
tick align=outside,
tick pos=left,
x grid style={darkgray176},
xmin=-0.5, xmax=811.5,
xtick style={color=black},
y dir=reverse,
y grid style={darkgray176},
ymin=-0.5, ymax=491.5,
ytick style={color=black},
]
\addplot graphics [includegraphics cmd=\pgfimage,xmin=-0.5, xmax=811.5, ymin=491.5, ymax=-0.5] {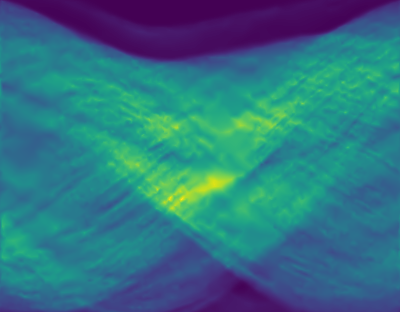};

\nextgroupplot[
tick align=outside,
tick pos=left,
x grid style={darkgray176},
xmin=-0.5, xmax=811.5,
xtick style={color=black},
y dir=reverse,
y grid style={darkgray176},
ymin=-0.5, ymax=491.5,
ytick style={color=black},
ylabel={\small Digimouse},
ylabel style={yshift=-\ys,},
]
\addplot graphics [includegraphics cmd=\pgfimage,xmin=-0.5, xmax=811.5, ymin=491.5, ymax=-0.5] {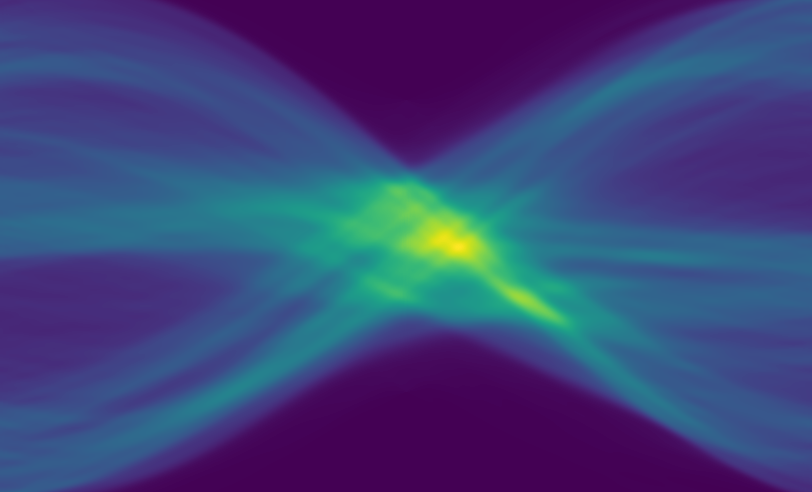};

\nextgroupplot[
hide x axis,
hide y axis,
tick align=outside,
tick pos=left,
x grid style={darkgray176},
xmin=-0.5, xmax=811.5,
xtick style={color=black},
y dir=reverse,
y grid style={darkgray176},
ymin=-0.5, ymax=491.5,
ytick style={color=black},
]
\addplot graphics [includegraphics cmd=\pgfimage,xmin=-0.5, xmax=811.5, ymin=491.5, ymax=-0.5] {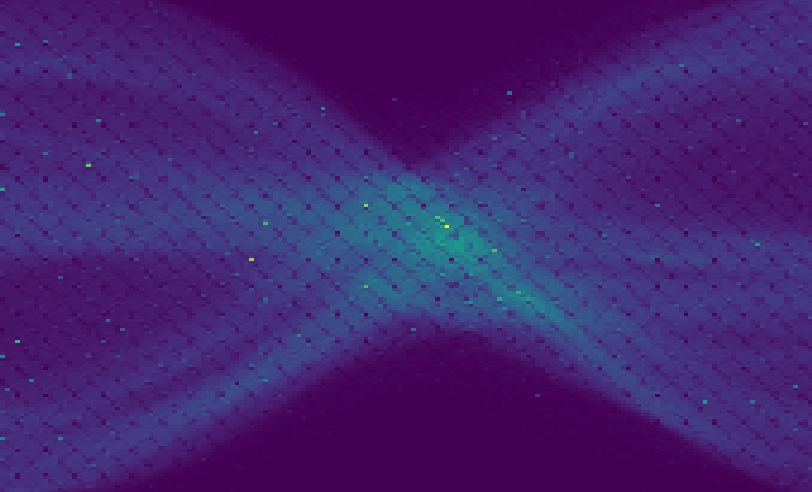};

\nextgroupplot[
hide x axis,
hide y axis,
tick align=outside,
tick pos=left,
x grid style={darkgray176},
xmin=-0.5, xmax=811.5,
xtick style={color=black},
y dir=reverse,
y grid style={darkgray176},
ymin=-0.5, ymax=491.5,
ytick style={color=black},
]
\addplot graphics [includegraphics cmd=\pgfimage,xmin=-0.5, xmax=811.5, ymin=491.5, ymax=-0.5] {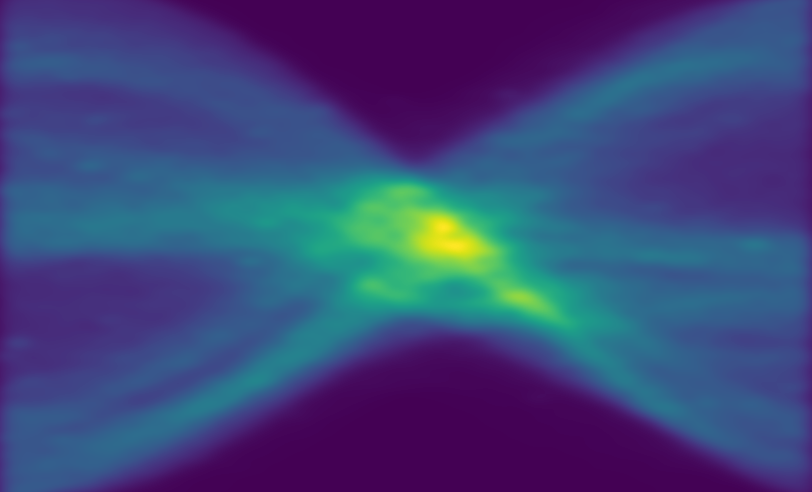};

\nextgroupplot[
hide x axis,
hide y axis,
point meta max=1.15928660492868e-05,
point meta min=1.15571540001439e-08,
tick align=outside,
tick pos=left,
x grid style={darkgray176},
xmin=-0.5, xmax=811.5,
xtick style={color=black},
y dir=reverse,
y grid style={darkgray176},
ymin=-0.5, ymax=491.5,
ytick style={color=black},
]
\addplot graphics [includegraphics cmd=\pgfimage,xmin=-0.5, xmax=811.5, ymin=491.5, ymax=-0.5] {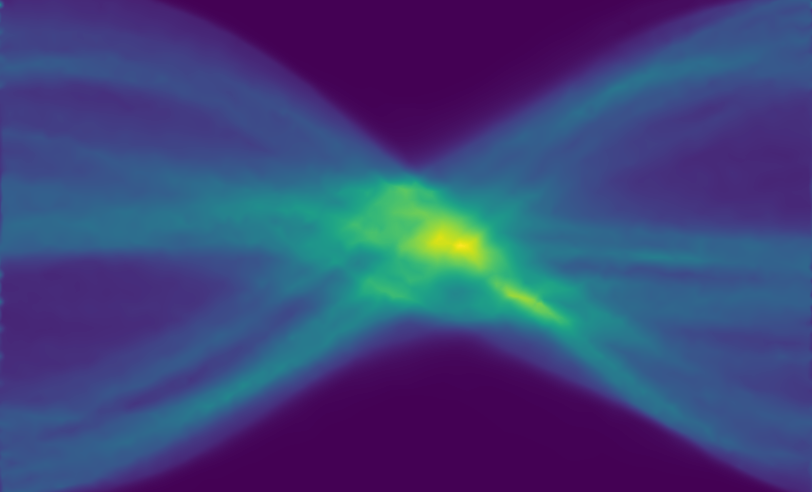};


\nextgroupplot[
tick align=outside,
tick pos=left,
x grid style={darkgray176},
xmin=-0.5, xmax=811.5,
xtick style={color=black},
y dir=reverse,
y grid style={darkgray176},
ymin=-0.5, ymax=491.5,
ytick style={color=black},
ylabel={\small Amyloid},
ylabel style={yshift=-\ys,},
]
\addplot graphics [includegraphics cmd=\pgfimage,xmin=-0.5, xmax=811.5, ymin=491.5, ymax=-0.5] {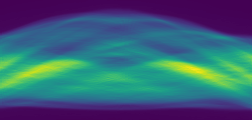};

\nextgroupplot[
hide x axis,
hide y axis,
tick align=outside,
tick pos=left,
x grid style={darkgray176},
xmin=-0.5, xmax=811.5,
xtick style={color=black},
y dir=reverse,
y grid style={darkgray176},
ymin=-0.5, ymax=491.5,
ytick style={color=black},
]
\addplot graphics [includegraphics cmd=\pgfimage,xmin=-0.5, xmax=811.5, ymin=491.5, ymax=-0.5] {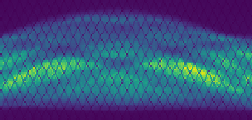};

\nextgroupplot[
hide x axis,
hide y axis,
tick align=outside,
tick pos=left,
x grid style={darkgray176},
xmin=-0.5, xmax=811.5,
xtick style={color=black},
y dir=reverse,
y grid style={darkgray176},
ymin=-0.5, ymax=491.5,
ytick style={color=black},
]
\addplot graphics [includegraphics cmd=\pgfimage,xmin=-0.5, xmax=811.5, ymin=491.5, ymax=-0.5] {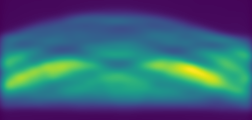};

\nextgroupplot[
hide x axis,
hide y axis,
point meta max=1.15928660492868e-05,
point meta min=1.15571540001439e-08,
tick align=outside,
tick pos=left,
x grid style={darkgray176},
xmin=-0.5, xmax=811.5,
xtick style={color=black},
y dir=reverse,
y grid style={darkgray176},
ymin=-0.5, ymax=491.5,
ytick style={color=black},
]
\addplot graphics [includegraphics cmd=\pgfimage,xmin=-0.5, xmax=811.5, ymin=491.5, ymax=-0.5] {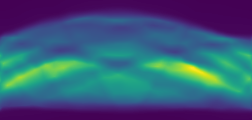};

\end{groupplot}

\end{tikzpicture}
    \vspace{-6pt}
    \caption{\tc{Sinograms estimated by the HE, KDE, and RDS. The corresponding MSEs~[dB] are (-118, -120, -127), (-117, -121, -126), (-119, -123, -128) for the Brain Phantom; (-124, -134, -141) for the Digimouse; and (-104, -107, -109) for the Amyloid. (See Appendix~D for the definition of the ground truth.)}}
    
    \label{fig:phantom_sinograms}
\end{figure*}

\begin{figure*}
    \centering
\begin{tikzpicture}

\definecolor{darkgray176}{RGB}{176,176,176}

\def\ys{-4pt}
\def\imwidth{.3\textwidth}\def\hs{1pt}\def\vs{1pt}
\begin{groupplot}[group style={group size=4 by 5,horizontal sep=\hs, vertical sep=\vs}, ticks=none, width=\imwidth]

\nextgroupplot[
tick align=outside,
tick pos=left,
x grid style={darkgray176},
xmin=-0.5, xmax=811.5,
xtick style={color=black},
y dir=reverse,
y grid style={darkgray176},
ymin=-0.5, ymax=491.5,
ytick style={color=black},
title={\small Ground Truth},
title style={yshift=\ys,},
ylabel={\small Brain Phantom},
ylabel style={yshift=-\ys,},
]
\addplot graphics [includegraphics cmd=\pgfimage,xmin=-0.5, xmax=811.5, ymin=491.5, ymax=-0.5] {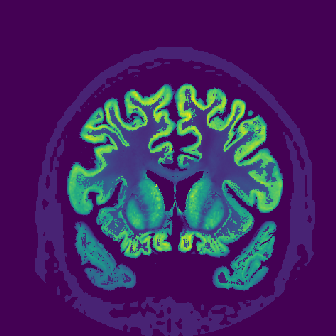};

\nextgroupplot[
hide x axis,
hide y axis,
tick align=outside,
tick pos=left,
x grid style={darkgray176},
xmin=-0.5, xmax=811.5,
xtick style={color=black},
y dir=reverse,
y grid style={darkgray176},
ymin=-0.5, ymax=491.5,
ytick style={color=black},
title={\small HE},
title style={yshift=\ys,},
]
\addplot graphics [includegraphics cmd=\pgfimage,xmin=-0.5, xmax=811.5, ymin=491.5, ymax=-0.5] {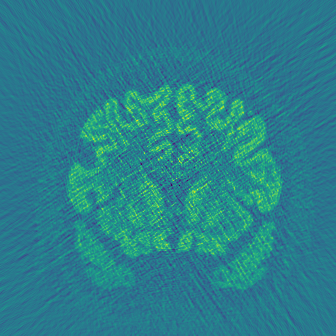};

\nextgroupplot[
hide x axis,
hide y axis,
tick align=outside,
tick pos=left,
x grid style={darkgray176},
xmin=-0.5, xmax=811.5,
xtick style={color=black},
y dir=reverse,
y grid style={darkgray176},
ymin=-0.5, ymax=491.5,
ytick style={color=black},
title={\small KDE},
title style={yshift=\ys,},
]
\addplot graphics [includegraphics cmd=\pgfimage,xmin=-0.5, xmax=811.5, ymin=491.5, ymax=-0.5] {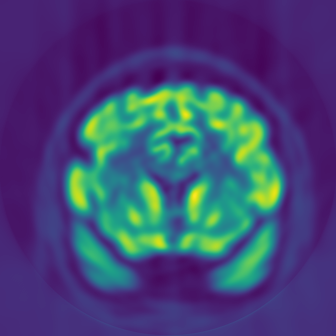};

\nextgroupplot[
hide x axis,
hide y axis,
point meta max=1.15928660492868e-05,
point meta min=1.15571540001439e-08,
tick align=outside,
tick pos=left,
x grid style={darkgray176},
xmin=-0.5, xmax=811.5,
xtick style={color=black},
y dir=reverse,
y grid style={darkgray176},
ymin=-0.5, ymax=491.5,
ytick style={color=black},
title={\small RDS},
title style={yshift=\ys,},
]
\addplot graphics [includegraphics cmd=\pgfimage,xmin=-0.5, xmax=811.5, ymin=491.5, ymax=-0.5] {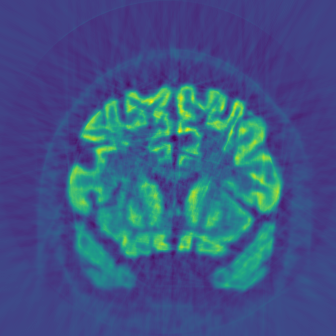};

\nextgroupplot[
tick align=outside,
tick pos=left,
x grid style={darkgray176},
xmin=-0.5, xmax=811.5,
xtick style={color=black},
y dir=reverse,
y grid style={darkgray176},
ymin=-0.5, ymax=491.5,
ytick style={color=black},
ylabel={\small Brain Phantom},
ylabel style={yshift=-\ys,},
]
\addplot graphics [includegraphics cmd=\pgfimage,xmin=-0.5, xmax=811.5, ymin=491.5, ymax=-0.5] {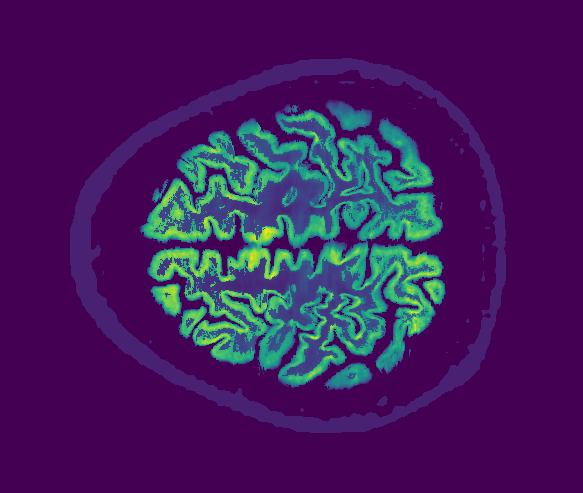};

\nextgroupplot[
hide x axis,
hide y axis,
tick align=outside,
tick pos=left,
x grid style={darkgray176},
xmin=-0.5, xmax=811.5,
xtick style={color=black},
y dir=reverse,
y grid style={darkgray176},
ymin=-0.5, ymax=491.5,
ytick style={color=black},
]
\addplot graphics [includegraphics cmd=\pgfimage,xmin=-0.5, xmax=811.5, ymin=491.5, ymax=-0.5] {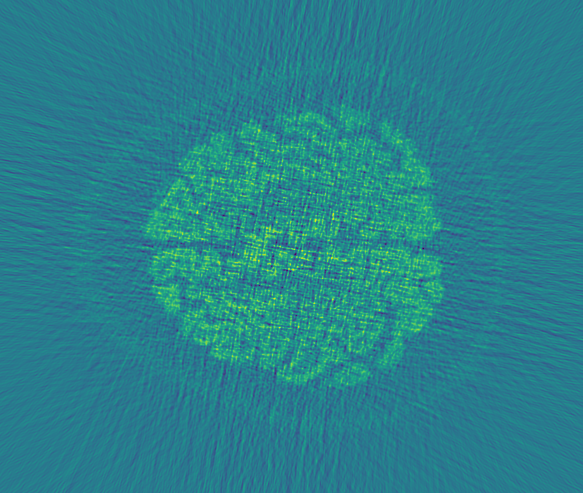};

\nextgroupplot[
hide x axis,
hide y axis,
tick align=outside,
tick pos=left,
x grid style={darkgray176},
xmin=-0.5, xmax=811.5,
xtick style={color=black},
y dir=reverse,
y grid style={darkgray176},
ymin=-0.5, ymax=491.5,
ytick style={color=black},
]
\addplot graphics [includegraphics cmd=\pgfimage,xmin=-0.5, xmax=811.5, ymin=491.5, ymax=-0.5] {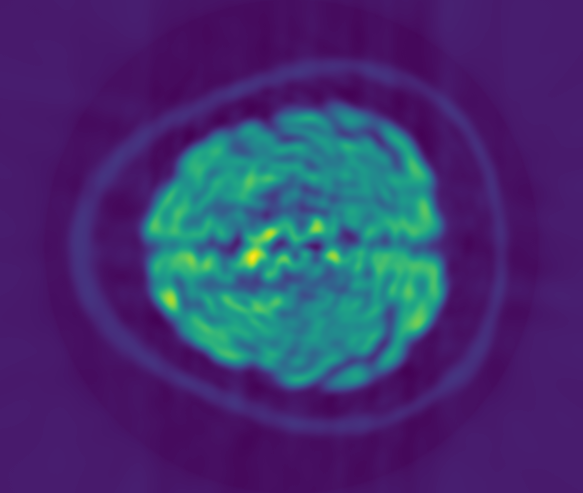};

\nextgroupplot[
hide x axis,
hide y axis,
point meta max=1.15928660492868e-05,
point meta min=1.15571540001439e-08,
tick align=outside,
tick pos=left,
x grid style={darkgray176},
xmin=-0.5, xmax=811.5,
xtick style={color=black},
y dir=reverse,
y grid style={darkgray176},
ymin=-0.5, ymax=491.5,
ytick style={color=black},
]
\addplot graphics [includegraphics cmd=\pgfimage,xmin=-0.5, xmax=811.5, ymin=491.5, ymax=-0.5] {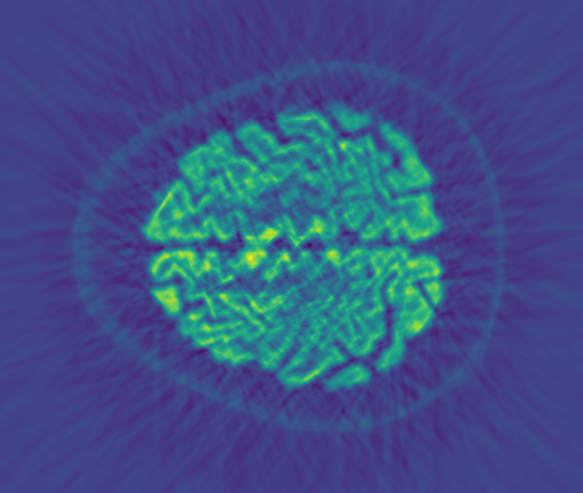};

\nextgroupplot[
tick align=outside,
tick pos=left,
x grid style={darkgray176},
xmin=-0.5, xmax=811.5,
xtick style={color=black},
y dir=reverse,
y grid style={darkgray176},
ymin=-0.5, ymax=491.5,
ytick style={color=black},
ylabel={\small Brain Phantom},
ylabel style={yshift=-\ys,},
]
\addplot graphics [includegraphics cmd=\pgfimage,xmin=-0.5, xmax=811.5, ymin=491.5, ymax=-0.5] {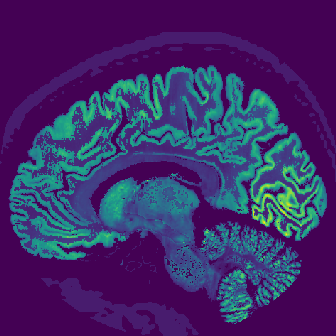};

\nextgroupplot[
hide x axis,
hide y axis,
tick align=outside,
tick pos=left,
x grid style={darkgray176},
xmin=-0.5, xmax=811.5,
xtick style={color=black},
y dir=reverse,
y grid style={darkgray176},
ymin=-0.5, ymax=491.5,
ytick style={color=black},
]
\addplot graphics [includegraphics cmd=\pgfimage,xmin=-0.5, xmax=811.5, ymin=491.5, ymax=-0.5] {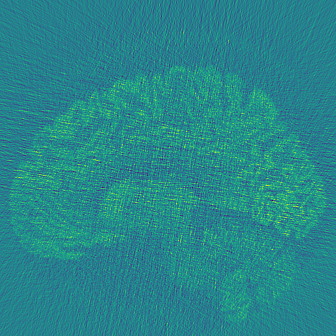};

\nextgroupplot[
hide x axis,
hide y axis,
tick align=outside,
tick pos=left,
x grid style={darkgray176},
xmin=-0.5, xmax=811.5,
xtick style={color=black},
y dir=reverse,
y grid style={darkgray176},
ymin=-0.5, ymax=491.5,
ytick style={color=black},
]
\addplot graphics [includegraphics cmd=\pgfimage,xmin=-0.5, xmax=811.5, ymin=491.5, ymax=-0.5] {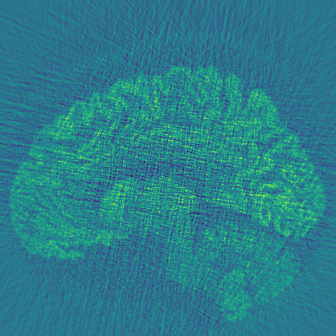};

\nextgroupplot[
hide x axis,
hide y axis,
point meta max=1.15928660492868e-05,
point meta min=1.15571540001439e-08,
tick align=outside,
tick pos=left,
x grid style={darkgray176},
xmin=-0.5, xmax=811.5,
xtick style={color=black},
y dir=reverse,
y grid style={darkgray176},
ymin=-0.5, ymax=491.5,
ytick style={color=black},
]
\addplot graphics [includegraphics cmd=\pgfimage,xmin=-0.5, xmax=811.5, ymin=491.5, ymax=-0.5] {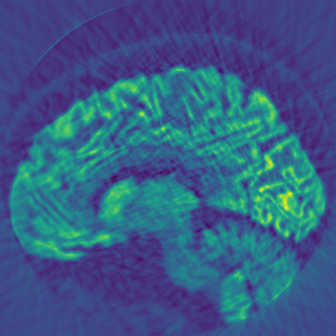};

\nextgroupplot[
tick align=outside,
tick pos=left,
x grid style={darkgray176},
xmin=-0.5, xmax=811.5,
xtick style={color=black},
y dir=reverse,
y grid style={darkgray176},
ymin=-0.5, ymax=491.5,
ytick style={color=black},
ylabel={\small Digimouse},
ylabel style={yshift=-\ys,},
]
\addplot graphics [includegraphics cmd=\pgfimage,xmin=-0.5, xmax=811.5, ymin=491.5, ymax=-0.5] {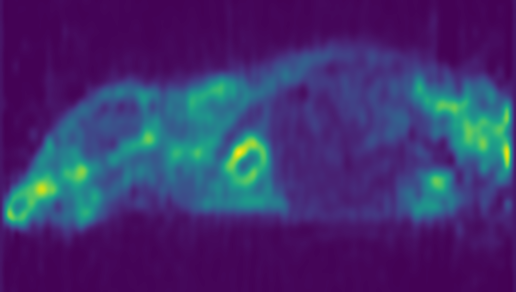};

\nextgroupplot[
hide x axis,
hide y axis,
tick align=outside,
tick pos=left,
x grid style={darkgray176},
xmin=-0.5, xmax=811.5,
xtick style={color=black},
y dir=reverse,
y grid style={darkgray176},
ymin=-0.5, ymax=491.5,
ytick style={color=black},
]
\addplot graphics [includegraphics cmd=\pgfimage,xmin=-0.5, xmax=811.5, ymin=491.5, ymax=-0.5] {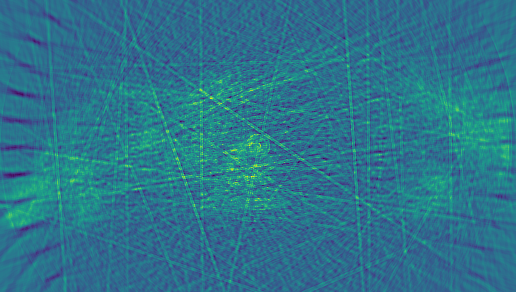};

\nextgroupplot[
hide x axis,
hide y axis,
tick align=outside,
tick pos=left,
x grid style={darkgray176},
xmin=-0.5, xmax=811.5,
xtick style={color=black},
y dir=reverse,
y grid style={darkgray176},
ymin=-0.5, ymax=491.5,
ytick style={color=black},
]
\addplot graphics [includegraphics cmd=\pgfimage,xmin=-0.5, xmax=811.5, ymin=491.5, ymax=-0.5] {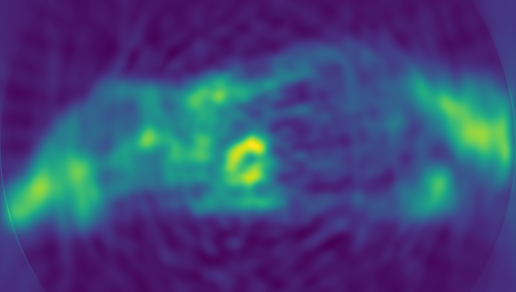};

\nextgroupplot[
hide x axis,
hide y axis,
point meta max=1.15928660492868e-05,
point meta min=1.15571540001439e-08,
tick align=outside,
tick pos=left,
x grid style={darkgray176},
xmin=-0.5, xmax=811.5,
xtick style={color=black},
y dir=reverse,
y grid style={darkgray176},
ymin=-0.5, ymax=491.5,
ytick style={color=black},
]
\addplot graphics [includegraphics cmd=\pgfimage,xmin=-0.5, xmax=811.5, ymin=491.5, ymax=-0.5] {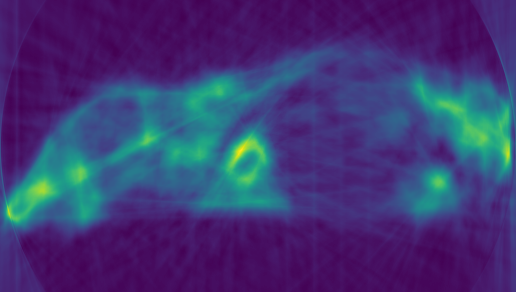};


\nextgroupplot[
tick align=outside,
tick pos=left,
x grid style={darkgray176},
xmin=-0.5, xmax=811.5,
xtick style={color=black},
y dir=reverse,
y grid style={darkgray176},
ymin=-0.5, ymax=491.5,
ytick style={color=black},
ylabel={\small Amyloid},
ylabel style={yshift=-\ys,},
]
\addplot graphics [includegraphics cmd=\pgfimage,xmin=-0.5, xmax=811.5, ymin=491.5, ymax=-0.5] {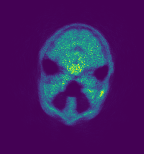};

\nextgroupplot[
hide x axis,
hide y axis,
tick align=outside,
tick pos=left,
x grid style={darkgray176},
xmin=-0.5, xmax=811.5,
xtick style={color=black},
y dir=reverse,
y grid style={darkgray176},
ymin=-0.5, ymax=491.5,
ytick style={color=black},
]
\addplot graphics [includegraphics cmd=\pgfimage,xmin=-0.5, xmax=811.5, ymin=491.5, ymax=-0.5] {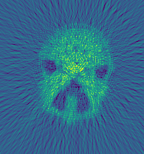};

\nextgroupplot[
hide x axis,
hide y axis,
tick align=outside,
tick pos=left,
x grid style={darkgray176},
xmin=-0.5, xmax=811.5,
xtick style={color=black},
y dir=reverse,
y grid style={darkgray176},
ymin=-0.5, ymax=491.5,
ytick style={color=black},
]
\addplot graphics [includegraphics cmd=\pgfimage,xmin=-0.5, xmax=811.5, ymin=491.5, ymax=-0.5] {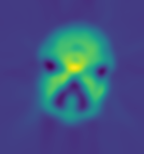};

\nextgroupplot[
hide x axis,
hide y axis,
point meta max=1.15928660492868e-05,
point meta min=1.15571540001439e-08,
tick align=outside,
tick pos=left,
x grid style={darkgray176},
xmin=-0.5, xmax=811.5,
xtick style={color=black},
y dir=reverse,
y grid style={darkgray176},
ymin=-0.5, ymax=491.5,
ytick style={color=black},
]
\addplot graphics [includegraphics cmd=\pgfimage,xmin=-0.5, xmax=811.5, ymin=491.5, ymax=-0.5] {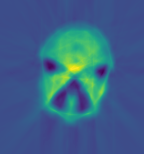};

\end{groupplot}

\end{tikzpicture}
    \vspace{-6pt}
    \caption{\tc{FBP reconstructions after resampling the sinograms with HE, KDE, and RDS. The corresponding MSEs~[dB] are (-225, -233, -240), (-236, -248, -254), (-225, -236, -239) for the Brain Phantom; (-226, -262, -269) for the Digimouse; and (-193, -200, -202) for the Amyloid.}}
    
    \label{fig:phantom_fbps}
\end{figure*}

\begin{figure*}
    \centering
\begin{tikzpicture}

\definecolor{darkgray176}{RGB}{176,176,176}

\def\ys{-4pt}
\def\imwidth{.3\textwidth}\def\hs{1pt}\def\vs{1pt}
\begin{groupplot}[group style={group size=4 by 5,horizontal sep=\hs, vertical sep=\vs}, ticks=none, width=\imwidth]

\nextgroupplot[
tick align=outside,
tick pos=left,
x grid style={darkgray176},
xmin=-0.5, xmax=811.5,
xtick style={color=black},
y dir=reverse,
y grid style={darkgray176},
ymin=-0.5, ymax=491.5,
ytick style={color=black},
title={\small Ground Truth},
title style={yshift=\ys,},
ylabel={\small Brain Phantom},
ylabel style={yshift=-\ys,},
]
\addplot graphics [includegraphics cmd=\pgfimage,xmin=-0.5, xmax=811.5, ymin=491.5, ymax=-0.5] {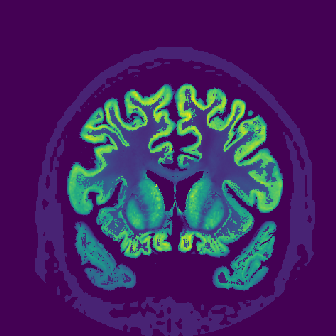};

\nextgroupplot[
hide x axis,
hide y axis,
tick align=outside,
tick pos=left,
x grid style={darkgray176},
xmin=-0.5, xmax=811.5,
xtick style={color=black},
y dir=reverse,
y grid style={darkgray176},
ymin=-0.5, ymax=491.5,
ytick style={color=black},
title={\small HE},
title style={yshift=\ys,},
]
\addplot graphics [includegraphics cmd=\pgfimage,xmin=-0.5, xmax=811.5, ymin=491.5, ymax=-0.5] {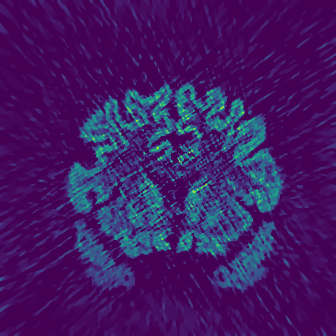};

\nextgroupplot[
hide x axis,
hide y axis,
tick align=outside,
tick pos=left,
x grid style={darkgray176},
xmin=-0.5, xmax=811.5,
xtick style={color=black},
y dir=reverse,
y grid style={darkgray176},
ymin=-0.5, ymax=491.5,
ytick style={color=black},
title={\small KDE},
title style={yshift=\ys,},
]
\addplot graphics [includegraphics cmd=\pgfimage,xmin=-0.5, xmax=811.5, ymin=491.5, ymax=-0.5] {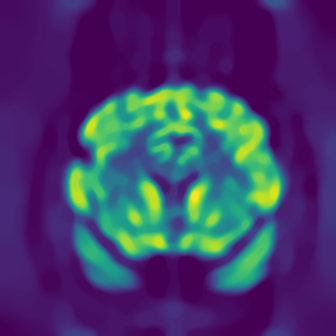};

\nextgroupplot[
hide x axis,
hide y axis,
point meta max=1.15928660492868e-05,
point meta min=1.15571540001439e-08,
tick align=outside,
tick pos=left,
x grid style={darkgray176},
xmin=-0.5, xmax=811.5,
xtick style={color=black},
y dir=reverse,
y grid style={darkgray176},
ymin=-0.5, ymax=491.5,
ytick style={color=black},
title={\small RDS},
title style={yshift=\ys,},
]
\addplot graphics [includegraphics cmd=\pgfimage,xmin=-0.5, xmax=811.5, ymin=491.5, ymax=-0.5] {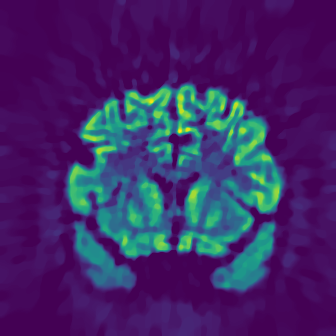};

\nextgroupplot[
tick align=outside,
tick pos=left,
x grid style={darkgray176},
xmin=-0.5, xmax=811.5,
xtick style={color=black},
y dir=reverse,
y grid style={darkgray176},
ymin=-0.5, ymax=491.5,
ytick style={color=black},
ylabel={\small Brain Phantom},
ylabel style={yshift=-\ys,},
]
\addplot graphics [includegraphics cmd=\pgfimage,xmin=-0.5, xmax=811.5, ymin=491.5, ymax=-0.5] {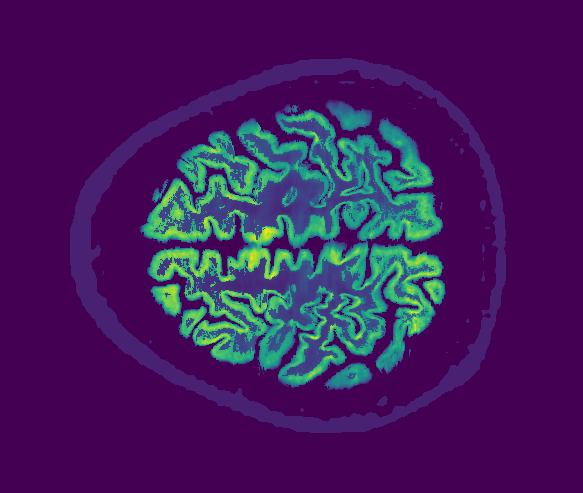};

\nextgroupplot[
hide x axis,
hide y axis,
tick align=outside,
tick pos=left,
x grid style={darkgray176},
xmin=-0.5, xmax=811.5,
xtick style={color=black},
y dir=reverse,
y grid style={darkgray176},
ymin=-0.5, ymax=491.5,
ytick style={color=black},
]
\addplot graphics [includegraphics cmd=\pgfimage,xmin=-0.5, xmax=811.5, ymin=491.5, ymax=-0.5] {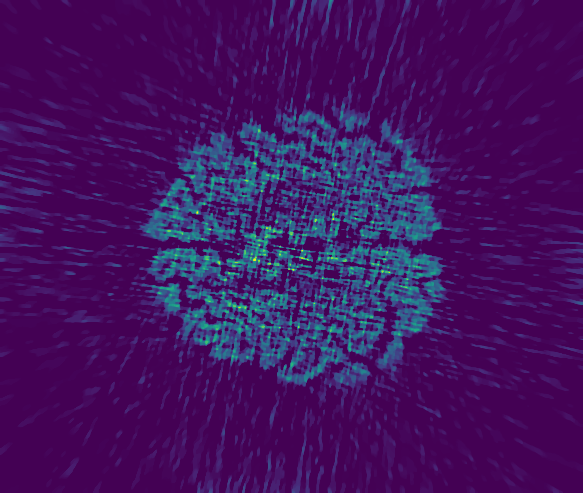};

\nextgroupplot[
hide x axis,
hide y axis,
tick align=outside,
tick pos=left,
x grid style={darkgray176},
xmin=-0.5, xmax=811.5,
xtick style={color=black},
y dir=reverse,
y grid style={darkgray176},
ymin=-0.5, ymax=491.5,
ytick style={color=black},
]
\addplot graphics [includegraphics cmd=\pgfimage,xmin=-0.5, xmax=811.5, ymin=491.5, ymax=-0.5] {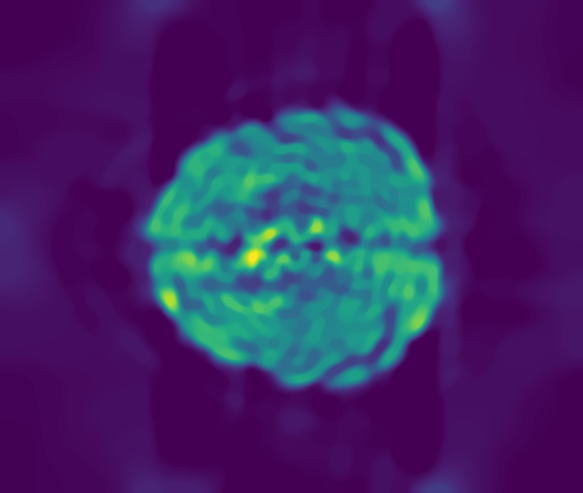};

\nextgroupplot[
hide x axis,
hide y axis,
point meta max=1.15928660492868e-05,
point meta min=1.15571540001439e-08,
tick align=outside,
tick pos=left,
x grid style={darkgray176},
xmin=-0.5, xmax=811.5,
xtick style={color=black},
y dir=reverse,
y grid style={darkgray176},
ymin=-0.5, ymax=491.5,
ytick style={color=black},
]
\addplot graphics [includegraphics cmd=\pgfimage,xmin=-0.5, xmax=811.5, ymin=491.5, ymax=-0.5] {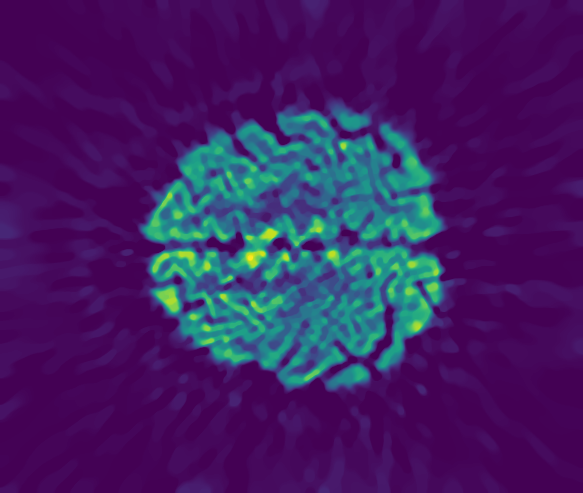};

\nextgroupplot[
tick align=outside,
tick pos=left,
x grid style={darkgray176},
xmin=-0.5, xmax=811.5,
xtick style={color=black},
y dir=reverse,
y grid style={darkgray176},
ymin=-0.5, ymax=491.5,
ytick style={color=black},
ylabel={\small Brain Phantom},
ylabel style={yshift=-\ys,},
]
\addplot graphics [includegraphics cmd=\pgfimage,xmin=-0.5, xmax=811.5, ymin=491.5, ymax=-0.5] {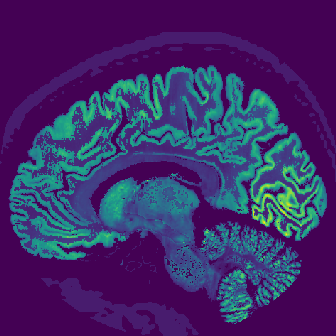};

\nextgroupplot[
hide x axis,
hide y axis,
tick align=outside,
tick pos=left,
x grid style={darkgray176},
xmin=-0.5, xmax=811.5,
xtick style={color=black},
y dir=reverse,
y grid style={darkgray176},
ymin=-0.5, ymax=491.5,
ytick style={color=black},
]
\addplot graphics [includegraphics cmd=\pgfimage,xmin=-0.5, xmax=811.5, ymin=491.5, ymax=-0.5] {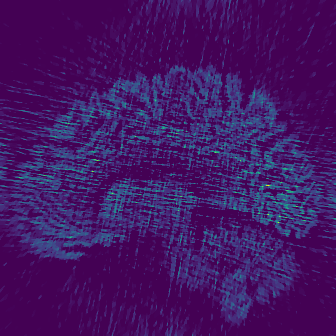};

\nextgroupplot[
hide x axis,
hide y axis,
tick align=outside,
tick pos=left,
x grid style={darkgray176},
xmin=-0.5, xmax=811.5,
xtick style={color=black},
y dir=reverse,
y grid style={darkgray176},
ymin=-0.5, ymax=491.5,
ytick style={color=black},
]
\addplot graphics [includegraphics cmd=\pgfimage,xmin=-0.5, xmax=811.5, ymin=491.5, ymax=-0.5] {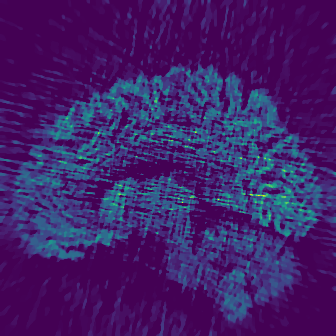};

\nextgroupplot[
hide x axis,
hide y axis,
point meta max=1.15928660492868e-05,
point meta min=1.15571540001439e-08,
tick align=outside,
tick pos=left,
x grid style={darkgray176},
xmin=-0.5, xmax=811.5,
xtick style={color=black},
y dir=reverse,
y grid style={darkgray176},
ymin=-0.5, ymax=491.5,
ytick style={color=black},
]
\addplot graphics [includegraphics cmd=\pgfimage,xmin=-0.5, xmax=811.5, ymin=491.5, ymax=-0.5] {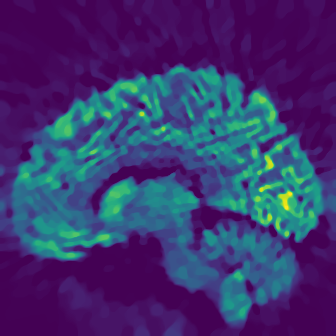};

\nextgroupplot[
tick align=outside,
tick pos=left,
x grid style={darkgray176},
xmin=-0.5, xmax=811.5,
xtick style={color=black},
y dir=reverse,
y grid style={darkgray176},
ymin=-0.5, ymax=491.5,
ytick style={color=black},
ylabel={\small Digimouse},
ylabel style={yshift=-\ys,},
]
\addplot graphics [includegraphics cmd=\pgfimage,xmin=-0.5, xmax=811.5, ymin=491.5, ymax=-0.5] {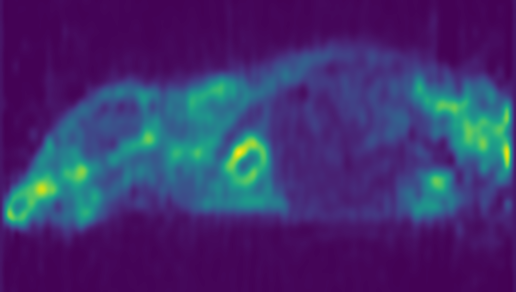};

\nextgroupplot[
hide x axis,
hide y axis,
tick align=outside,
tick pos=left,
x grid style={darkgray176},
xmin=-0.5, xmax=811.5,
xtick style={color=black},
y dir=reverse,
y grid style={darkgray176},
ymin=-0.5, ymax=491.5,
ytick style={color=black},
]
\addplot graphics [includegraphics cmd=\pgfimage,xmin=-0.5, xmax=811.5, ymin=491.5, ymax=-0.5] {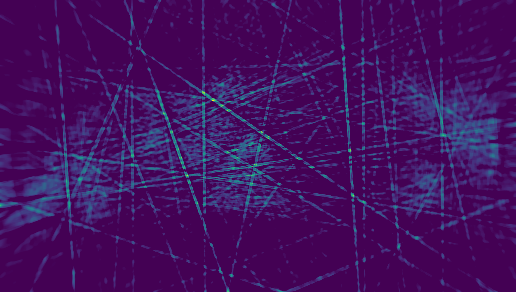};

\nextgroupplot[
hide x axis,
hide y axis,
tick align=outside,
tick pos=left,
x grid style={darkgray176},
xmin=-0.5, xmax=811.5,
xtick style={color=black},
y dir=reverse,
y grid style={darkgray176},
ymin=-0.5, ymax=491.5,
ytick style={color=black},
]
\addplot graphics [includegraphics cmd=\pgfimage,xmin=-0.5, xmax=811.5, ymin=491.5, ymax=-0.5] {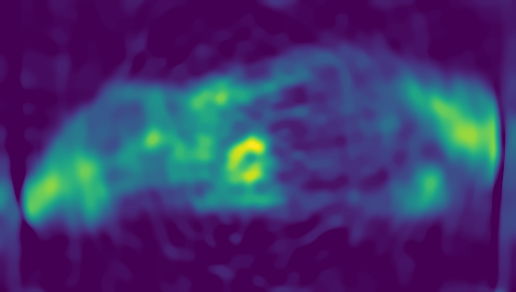};

\nextgroupplot[
hide x axis,
hide y axis,
point meta max=1.15928660492868e-05,
point meta min=1.15571540001439e-08,
tick align=outside,
tick pos=left,
x grid style={darkgray176},
xmin=-0.5, xmax=811.5,
xtick style={color=black},
y dir=reverse,
y grid style={darkgray176},
ymin=-0.5, ymax=491.5,
ytick style={color=black},
]
\addplot graphics [includegraphics cmd=\pgfimage,xmin=-0.5, xmax=811.5, ymin=491.5, ymax=-0.5] {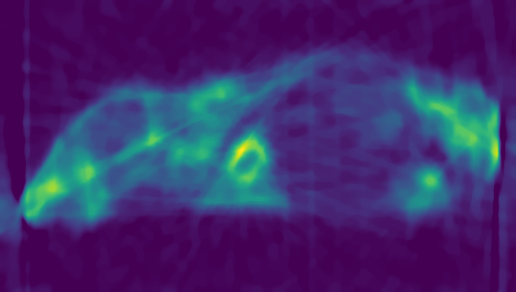};


\nextgroupplot[
tick align=outside,
tick pos=left,
x grid style={darkgray176},
xmin=-0.5, xmax=811.5,
xtick style={color=black},
y dir=reverse,
y grid style={darkgray176},
ymin=-0.5, ymax=491.5,
ytick style={color=black},
ylabel={\small Amyloid},
ylabel style={yshift=-\ys,},
]
\addplot graphics [includegraphics cmd=\pgfimage,xmin=-0.5, xmax=811.5, ymin=491.5, ymax=-0.5] {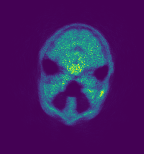};

\nextgroupplot[
hide x axis,
hide y axis,
tick align=outside,
tick pos=left,
x grid style={darkgray176},
xmin=-0.5, xmax=811.5,
xtick style={color=black},
y dir=reverse,
y grid style={darkgray176},
ymin=-0.5, ymax=491.5,
ytick style={color=black},
]
\addplot graphics [includegraphics cmd=\pgfimage,xmin=-0.5, xmax=811.5, ymin=491.5, ymax=-0.5] {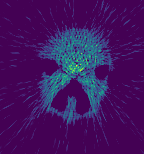};

\nextgroupplot[
hide x axis,
hide y axis,
tick align=outside,
tick pos=left,
x grid style={darkgray176},
xmin=-0.5, xmax=811.5,
xtick style={color=black},
y dir=reverse,
y grid style={darkgray176},
ymin=-0.5, ymax=491.5,
ytick style={color=black},
]
\addplot graphics [includegraphics cmd=\pgfimage,xmin=-0.5, xmax=811.5, ymin=491.5, ymax=-0.5] {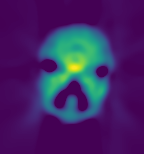};

\nextgroupplot[
hide x axis,
hide y axis,
point meta max=1.15928660492868e-05,
point meta min=1.15571540001439e-08,
tick align=outside,
tick pos=left,
x grid style={darkgray176},
xmin=-0.5, xmax=811.5,
xtick style={color=black},
y dir=reverse,
y grid style={darkgray176},
ymin=-0.5, ymax=491.5,
ytick style={color=black},
]
\addplot graphics [includegraphics cmd=\pgfimage,xmin=-0.5, xmax=811.5, ymin=491.5, ymax=-0.5] {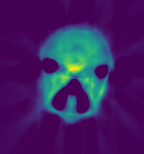};

\end{groupplot}

\end{tikzpicture}
    \vspace{-6pt}
    \caption{\tc{Total-variation reconstructions after resampling the sinograms with HE, KDE, and RDS. The corresponding MSEs~[dB] are (-233, -233, -241), (-245, -249, -255), (-234, -236, -240) for the Brain Phantom; (-228, -262, -269) for the Digimouse; and (-193, -200, -203) for the Amyloid.}}
    
    \label{fig:phantom_tvs}
\end{figure*}

\tc{We first used the ultrahigh-resolution PET phantom proposed by \cite{belzunce_technical_2020} (Brain Phantom). We simulated the emission and acquisition thereof in the PET scanner of~\cite{cadoux_100mupet_2023}. (See Appendix D for more details and for a comment about simulations in PET.) We resampled the sinogram using the three DE methods as described above. The PET images were then reconstructed from the resulting sinograms with the FBP (as per the NEMA standard). We also reconstructed the PET images using total-variation regularization. The resampled sinograms and their corresponding reconstructions are shown in Figures~\ref{fig:phantom_sinograms} and~\ref{fig:phantom_fbps}-\ref{fig:phantom_tvs}, respectively.}

\tc{We also tested our method on the Digimouse~\cite{dogdas_digimouse_2007} as well as on the Amyloid real dataset, which is a scan acquired with the widespread PET-MR scanner Siemens Biograph mMR 3T~\cite{lane_study_2017,markiewicz_single_2018} (see Appendix D). Their sinograms and reconstructions are in Figures~\ref{fig:phantom_sinograms}-~\ref{fig:phantom_tvs}, too.
}

\tc{Our observations are similar for all three experiments (Brain Phantom, Digimouse, Amyloid). Zones of low sensitivity in the sinogram are problematic for HE, partly because of the stability issues introduced by the sensitivity inversion. In these situations, KDE oversmooths the image, in an attempt to compensate for the low sensitivity. This overlooks potential sharp zones that might exist in the underlying sinogram. This issue is compounded when the regions of low sensitivity (or with low sinogram values) have different scales or sharpness. For example, see the grid of ``gap'' zones in the amyloid dataset or in the Digimouse. The RDS fares comparatively well against the heterogeneity of both the underlying sinogram and the sensitivity. This is noticeable in the present case of tomographic imaging, even if the x-ray transform (which leads to the sinograms) is smoothing~\cite{markoe_analytic_2006,parhi_distributional_2023}.
In addition to providing sharper structures, the RDS also reproduces the original contrast better. This is especially important in PET because the contrast delineates the regions where the uptake of the radiotracer differs. Similar effects are visible in both the simulated and the real experiments.} \tc{Upon reconstruction using total-variation regularization (as compared to FBP) all three methods gained better contrast with respect to the background (compare Figures~\ref{fig:phantom_fbps} and \ref{fig:phantom_tvs}).
}

\section{Conclusion and Discussion}
Regularized-density splines (RDS) compensate for heterogeneous sensitivities and scale well with dimension. The approach is robust to the choice of regularization parameter. This is a good asset for density estimation (DE) because it plays the role of bandwidth selection. The role of the regularization is also well substantiated: it induces Hessian sparsity under an invariant framework. Optimization and evaluation times are independent of the number of samples, with evaluations being completely parallelizable. An implementation of RDS is available via a GPU-supported library.

All these characteristics make RDS a good candidate for several applications, especially imaging ones. We validated that modern PET scanners with small detectors can be approached via \de in general, and that RDS was a particularly good choice. This approach is also applicable to low-resolution scanners if quantization were accounted for \cite{del_aguila_pla_convex_2022}. With nanotechnology fostering ever smaller detectors, we expect that other imaging modalities will enter the regime of weighted statistical sampling. This is because localization is growing more accurate, but signal power is likely to remain constant---limited by dose or exposure.

Future directions will consist in futher investigations of the link between the regularization term and the sparsity of splines. Box splines hold great potential in this direction because they are piecewise linear, but they may come at increased computational and implementation costs. Other efforts will focus on the tailoring of data terms to image reconstruction based on the uncertainty of RDS estimates.

\section*{Acknowledgments}
We would like to thank the anonymous reviewers and the editors for their efforts and for their valuable feedback. 
We are grateful to the team of Prof.~Giuseppe Iacobucci at University of Geneva for the specifications of the PET scanner. 
We  also want to acknowledge Thomas Debarre for helpful discussions about splines. 
This work was funded (in part) by the Swiss National Science Foundation under the Sinergia 
grant CRSII5\_198569. 
We acknowledge access to the facilities and expertise of the CIBM Center
for Biomedical Imaging, a Swiss research center of excellence founded and
supported by Lausanne University Hospital (CHUV), University of Lausanne
(UNIL), École polytechnique fédérale de Lausanne (EPFL), University of
Geneva (UNIGE), and Geneva University Hospitals (HUG). A.B.-P. conceived the project, developed the framework, wrote the code, and prepared the manuscript. P.d.A.P contributed with discussions about 
the design of the experiments and about the state of the art of DE. 
M.U. revised the work and contributed discussions about splines. All the authors discussed the results, and reviewed and approved the manuscript. The authors declare no competing interests.

\ifCLASSOPTIONcaptionsoff
  \newpage
\fi

\bibliographystyle{IEEEtran}
\bibliography{IEEEabrv,sample,references}

\begin{IEEEbiography}[{\includegraphics[width=1in,height=1.25in,clip,keepaspectratio]{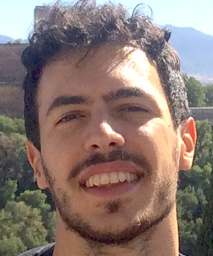}}]{Aleix Boquet-Pujadas} received a bachelor's degree in Mathematics and a bachelor's degree in Physics from Universitat Aut\`onoma de Barcelona (first-in-class) and now holds a Ph.D. from Sorbonne Universit\'e (or UPMC, or Paris VI). His doctoral thesis was supported by a Marie Sk$\l{}$odowska-Curie fellowship at Institut Pasteur and distinguished by the French Society of Biomedical Engineering (SFGBM) for its innovation. He is currently a postdoctoral researcher at the Biomedical Imaging Group, EPFL.
\end{IEEEbiography}

\begin{IEEEbiography}[{\includegraphics[width=1in,height=1.25in,clip,keepaspectratio]{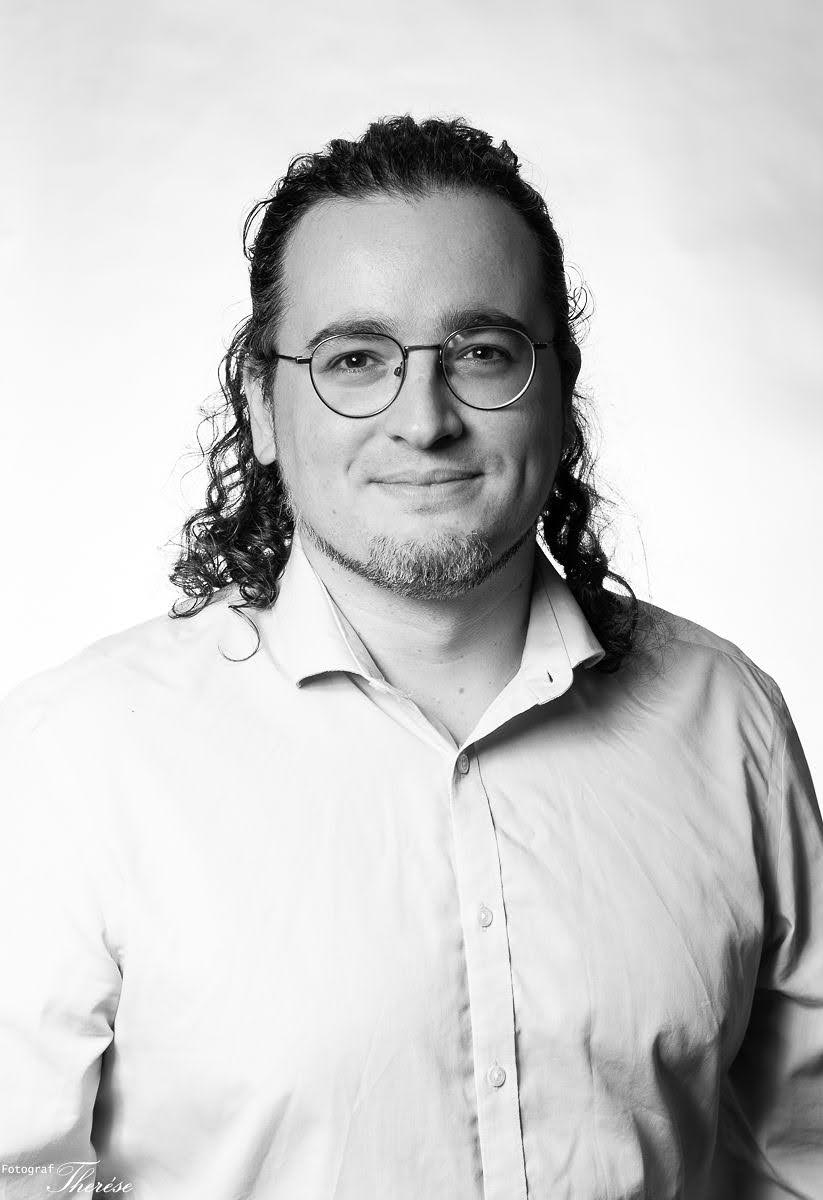}}]{Pol del Aguila Pla} (S’15-M’19) received a double degree in telecommunications and electrical engineering from Universitat Politècnica de Catalunya (UPC) and KTH Royal Institute of Technology in 2014, and a Ph.D. in electrical engineering from KTH Royal Institute of Technology in 2019. After a postdoctoral at the CIBM Center for Biomedical Imaging (CIBM) in Switzerland, at the EPFL’s Biomedical Imaging Group in Lausanne, Switzerland, he now works as an image processing specialist at Qamcom Research and Technology in Stockholm, Sweden. 

Pol’s research interest cover advanced uses of signal processing and applied mathematics across the engineering sciences.
\end{IEEEbiography}

\begin{IEEEbiography}[{\includegraphics[width=1in,height=1.25in,clip,keepaspectratio]{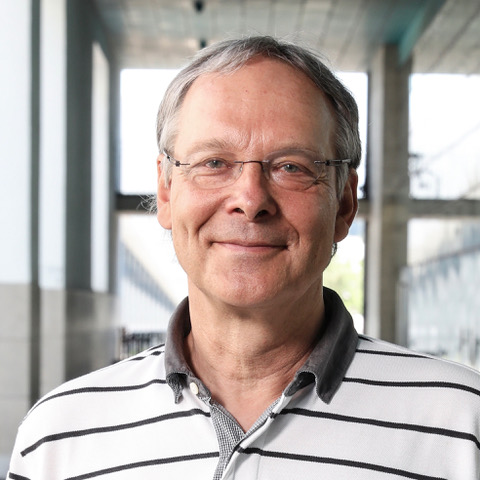}}]{Michael Unser} (M’89–SM’94–F’99–LF’23) is Full Professor at the EPFL and the academic director of EPFL's Center for Imaging, Lausanne, Switzerland.
His primary areas of investigation are biomedical imaging and applied functional analysis. He is internationally recognized for his research contributions to sampling theory, wavelets, the use of splines for image processing, stochastic processes, and computational bioimaging. He has published over 400 journal papers on those topics. He is the author with P. Tafti of the book “An introduction to sparse stochastic processes”, Cambridge University Press 2014. 

From 1985 to 1997, he was with the Biomedical Engineering and Instrumentation Program, National Institutes of Health, Bethesda USA, conducting research on bioimaging. Dr. Unser has served on the editorial board of most of the primary journals in his field including the IEEE Transactions on Medical Imaging (associate Editor-in-Chief 2003-2005), IEEE Trans. Image Processing, Proc. of IEEE, and SIAM J. of Imaging Sciences. He co-organized the first IEEE International Symposium on Biomedical Imaging (ISBI2002) and was the founding chair of the technical committee of the IEEE-SP Society on Bio Imaging and Signal Processing (BISP).

Prof. Unser is a fellow of the IEEE (1999), an EURASIP fellow (2009), and a member of the Swiss Academy of Engineering Sciences. He is the recipient of several international prizes including five IEEE-SPS Best Paper Awards, two Technical Achievement Awards from the IEEE (2008 SPS and EMBS 2010), the 2018 Technical Achievement Award from EURASIP, and the 2020 Career Achievement Award from the IEEE Society on Engineering in Medicine and Biology (EMBS). He was awarded three ERC AdG grants: FUNSP (2011-2016), GlobalBioIm (2016-2021), and FunLearn (2021-2026).
\end{IEEEbiography}

\end{document}


\bstctlcite{IEEEexample:BSTcontrol}

\title{Sensitivity-Aware Density Estimation in Multiple Dimensions} 

\author{Aleix Boquet-Pujadas
\thanks{Corresponding author: aleix.boquetipujadas@epfl.ch\\
The authors are with the Biomedical Imaging Group at the École polytechnique fédérale de Lausanne, 
Lausanne, Switzerland.}, Pol {del Aguila Pla},~\IEEEmembership{Member,~IEEE,}%
\thanks{Pol~{del Aguila Pla} is also with the CIBM Center for Biomedical Imaging, in Switzerland.}
and Michael Unser,~\IEEEmembership{Fellow,~IEEE}%
\thanks{Manuscript received MMMM DD, YYYY; revised MMMM DD, YYYY.}}


\maketitle

\IEEEraisesectionheading{\textbf{APPENDICES}}
\appendices
\section{Connection Between Regularization and Free Knots}
\label{appendix:free_knots}
\tc{In standard logsplines on the real line, one places the knots of the spline according to a number of rules of thumb that may involve placing knots near order statistics or placing middle knots at equally spaced intervals~[29]
The number of knots can be chosen according to experimental tables. At first, the spline coefficients are fit to the data via optimization. In a second (optional) step, the number of knots can then be progressively reduced. If this second step is used, one chooses a large number of initial knots. The knots are then deleted one at a time in a greedy fashion by eliminating the one that (locally) contributes the least information according to some criterium.}

\tc{Our regularization acts somewhat similarly to this concept of free-knot logsplines. It ``deactivates'' knots in an automatic and principled manner, following a general criterium. It does so on multidimensional compact domains and is single-step, convex, and based on convolutions. The explanation for the deactivation effect is as follows.}

\tc{\textbullet \, Set up a linear B-spline expansion on a fine, uniform grid with (too) many knots. We remind that operations are more efficient therewith because of the uniformity, and that B-splines are basis functions for splines with the same knots.}

\tc{\textbullet \, Consider the following result in dimension $d=1$. Let $M$ be the space of bounded Radon measures equipped with the total-variation norm for measures. Then, when combined with a data-fidelity cost term, the regularization term 
\begin{equation}\label{eq:mnorm}
\lambda||D^2 f||_M    
\end{equation}
favors solutions that are nonuniform linear splines with few knots~\cite{debarre_sparsest_2022}. (This result does not assume any parameterization of the function $f:\reals \to \reals$.) Here, $D^2$ is the second-derivative operator. The number of knots of the solution spline tends to $0$ as $\lambda$ grows, as sparsity is then promoted more strongly.}

\tc{\textbullet \, For $d=1$, our proposed regularization on the Hessian of the logarithm 
(20) 
is equivalent to the discretization of the regularization \eqref{eq:mnorm} above. In this theoretical context, our function $f$ (the logarithm of the density) is pre-parameterized by the uniform B-spline expansion, which can approximate the expected solutions very well if the grid is fine enough. More intuitively, the least relevant knots of the linear B-spline will be deactivated by the regularization: The minimization of the second derivative will make the function locally linear (affine, in fact), effectively making it look as if some knots were ``skipped'' or ``eliminated''. This effect can be seen in Figure~2-top 
(slice of $d=2$). The nonuniform piecewise-linear regions of the solution are expressible by the finer piecewise-linear regions of the parameterization. In other words, the initial uniform grid with superabundant knots turns into one that appears nonuniform. This nonuniformity is chosen in a principled manner that promotes~sparsity.}

\tc{The argument generalizes to higher dimensions because the combination of the Hessian with the total-variation favors piecewise-linear solutions~\cite{aziznejad_measuring_2023,ambrosio_linear_2023,ambrosio_functions_2023} that are locally affine. It promotes solutions with fewer facets or linear regions. Moreover, our choice of nuclear norm acts as a surrogate for the promotion of low-rank (Hessian, in our case) matrices.}



\section{Convolutions for Normalization Integrals} 
\label{appendix:normalization_integral}
\def\a{a}
\def\veps{\sensi_s}
\def\nonoutm{\mathrm{\mathbf{D}}}

The coupling of $\densitye$ and $\sensi$ via their normalization in  
(3) 
comes at the price of an integration step. Such steps are generally expensive but, here, the cost is affordable because the dimension of the domain is low, the B-spline basis has a small support, and the exponential of the B-spline is smooth. To ease the computational burden even further, we propose to exploit the convolutional properties of B-spline evaluations on a uniform grid. 
We formulate
\begin{equation}
\a_s \define \veps \exp \left( \ker^{(\mathbf{0})}_s * \coeffs_{\uparrow s }\right)
\end{equation}
to approximate the integral $\Sensi(\densitye)$ as
\begin{equation}
\Sensi(\densitye) \approx s^{-d} \norm{\a_s}_1,
\end{equation}
where the arrow refers to upsampling with proper insertion of zeros, and $\veps[\mathbf{k}]$ is the function $\sensi$ sampled at scale $s$ in consideration of $\vmu$, and that can either be precomputed or rewritten in a basis of logsplines. We remark that, for linear splines, the integral has a closed form in one dimension. When the domain is multidimensional, it reduces to a one-dimensional integral. 

Similarly, we can write the term $\Sensi(\vbeta_\k \densitye)$ in $\mathbf{g}$ and $\mathbf{f}$ with a separable convolution as
\begin{equation}
\Sensi(\vbeta_\k \densitye) \approx s^{-d} \left(\ker^{(\mathbf{0})}_s * \a_s \right)_{\downarrow s } [\k].
\end{equation}
The terms $\Sensi(\vbeta_\k \vbeta_{\k-\m} \densitye)$ in $\nonoutm$ are
\def\bb{\ff}
\begin{equation}
\Sensi(\vbeta_\k \vbeta_{\k-\m} \densitye)\approx \left(\, {\bb}_{s, \k-\m} * \a_s \right)_{\downarrow s }
[\k, \k-\m]
\end{equation}
for $\norm{\k-\m}_\infty<s(n+1)$ and $0$ otherwise, where the convolution is separable but needs to be applied once per filter. The filter $\bb_{s, \k-\m}$ is given by the outer product between the one-dimensional filters stemming from all nonzero overlaps
$\beta_s(\cdot-0)\beta_s(\cdot-(k-m))$
between pairs of B-splines of degree $n$; there are $(s(n+1)-1)$ different one-dimensional filters with support of size $(s(n+1)-1)$ when accounting for symmetries.


\section{Efficient Evaluation of the Target Density}
\label{appendix:evaluation_density}
The optimal coefficients $\opticoeffs$ in 
(16) 
lead to an estimate $\densitye (\opticoeffs)$ for the unnormalized target density. An estimate for the density of interest $\density(\point)$ is therefore
\begin{equation}
 \hat{\density}(\point) =  \frac{\densitye (\point; \opticoeffs) }{ \int_{\sinodomain} \densitye}.
\end{equation}
In practice, we compute the normalization constant $\int_{\sinodomain} \densitye$ by approximating it at a reference scale $h$ as
\begin{equation}
I_h \define h^{-d}\norm{\exp{\left( \ker^{(\mathbf{0})}_h * \opticoeffs_{\uparrow h }\right)}}_1.
\end{equation}
The evaluation $\samp_{\mathcal{M}_s}\{ \density\}$ of the density can then be evaluated on a grid $\mathcal{M}_s$ of points at some scale $s<h$ with respect to the original grid $\mathcal{M}$ as
\begin{equation}
I_h^{-1} \samp_{\mathcal{M}_s}\{  \densitye\}
\end{equation}
through a simple convolution
\begin{equation}\label{eq:convolution_evaluation}
\samp_{\mathcal{M}_s}\{  \densitye\}
= \exp{\left( \ker^{(\mathbf{0})}_s * \coeffs_{\uparrow s }^\star \right)}.
\end{equation}
Evaluations on subgrids such as \eqref{eq:convolution_evaluation} have a very low complexity. While not quite as efficient, evaluations on irregular sets can benefit from the compact support of splines to greatly reduce the number of computations.

\section{PET Data and Phantoms}
\label{appendix:pet_data}
\tc{The Brain Phantom~[51] 
combines the resolution and heterogeneity of the BigBrain atlas with the activity levels available in the Hammersmith atlas. It is based on an acquisition using a Siemens mMR scanner of a female epilepsy patient with a  [$^{18}$F]fluorodeoxyglucose (FDG) radiotracer. Its purpose is to reproduce the heterogeneous uptake in the brain (observable with PET) more accurately than the pioneering BrainWeb phantom~\cite{collins_design_1998}, which is based on (and preeminently designed for) an MRI data set. The data were scaled to fit inside the simulated scanner~[49]
. }

\tc{We remark that simulations are widespread in the field of PET because they are able to model most physical effects known to play a role in the emission and acquisition processes. 
They have been shown to produce results that are very close to those obtained in real experiments~\cite{lu_validation_2016, ahmed_validated_2020}, both in terms of figures of merit (e.g., resolution or sensitivity) and reconstruction artifacts.
}

\tc{The Digimouse was obtained by imaging an entire frozen mouse injected with FDG ($765$ $\mu$Ci) and $^{18}$F$^{-}$ ($216$ $\mu$Ci) in a Concorde P4 microPET scanner for $30$ minutes after $60$ minutes of uptake~[52].
}

\tc{The Amyloid PET data were collected by imaging the brain of a human patient injected with the amyloid ligand $^{18}$F-florbetapir ($370$ MBq) in a Siemens Biograph mMR $3$T for $60$ minutes~[53],[54]. 
To have a ``groundruth'', we subsampled the data. In particular, we used only the last $10$ minutes, which allows time for uptake and, thus, for an estimate of the amyloid load. The MSEs must, therefore, not be taken at face value since they are computed with respect to the full-time scan. Only the normalization and patient's attenuation were weighted because the hardware attenuation map is proprietary. We believe this to be the cause of the streaks (not the grid-patterned ``gaps'') observed in the Amyloid sinograms upon resampling with the three methods. 
}




\let\oldthebibliography=\thebibliography
\let\oldendthebibliography=\endthebibliography
\renewenvironment{thebibliography}[1]{
    \oldthebibliography{#1}
    \setcounter{enumiv}{57}                        
}{\oldendthebibliography}
\bibliographystyle{IEEEtran}
\bibliography{IEEEabrv,sample,references}
\setcounter{enumiv}{8}